\documentclass{article}

\usepackage[preprint,nonatbib]{neurips_2023}

\usepackage[utf8]{inputenc} 
\usepackage[T1]{fontenc}    
\usepackage[colorlinks,linkcolor=blue]{hyperref} 
\usepackage{url}            
\usepackage{booktabs}       
\usepackage{amsfonts}       
\usepackage{nicefrac}       
\usepackage{microtype}      
\usepackage{xcolor}         

\usepackage{amssymb}
\usepackage{amsmath}
\usepackage{graphicx}
\usepackage{subcaption}
\usepackage{multirow}
\usepackage{CJKutf8}
\usepackage{tabularx}
\usepackage{adjustbox}
\usepackage[misc]{ifsym}
\usepackage{wrapfig}
\usepackage[linesnumbered,ruled,vlined]{algorithm2e}

\usepackage{enumitem}
\setlist[itemize]{leftmargin=*}

\title{GlyphDraw: Seamlessly Rendering Text with Intricate Spatial Structures in Text-to-Image Generation}

\author{%
  Jian Ma \\
  OPPO Research Institute \\
  \texttt{majian2@oppo.com} \\
  \And
  Mingjun Zhao \footnotemark[1] \\
  University of Alberta \\
  \texttt{zhao2@ualberta.ca} \\
  \AND
  Chen Chen$^{\textrm{\Letter}}$ \\
  OPPO Research Institute  \\
  \texttt{chenchen4@oppo.com} \\
  \And
  Ruichen Wang \\
  OPPO Research Institute \\
  \texttt{wangruichen@oppo.com} \\
  \And
  Di Niu \\
  University of Alberta \\
  \texttt{dniu@ualberta.ca} \\
  \And
  Haonan Lu$^{\textrm{\Letter}}$ \\
  OPPO Research Institute \\
  \texttt{luhaonan@oppo.com} \\
  \And
  Xiaodong Lin \\
  Rutgers University \\
  \texttt{lin@business.rutgers.edu} \\
}

\begin{document}

\maketitle

\footnotetext[1]{Author did this work during his internship at OPPO Research Institute.}

\begin{figure}[h]
\includegraphics[width=\linewidth]{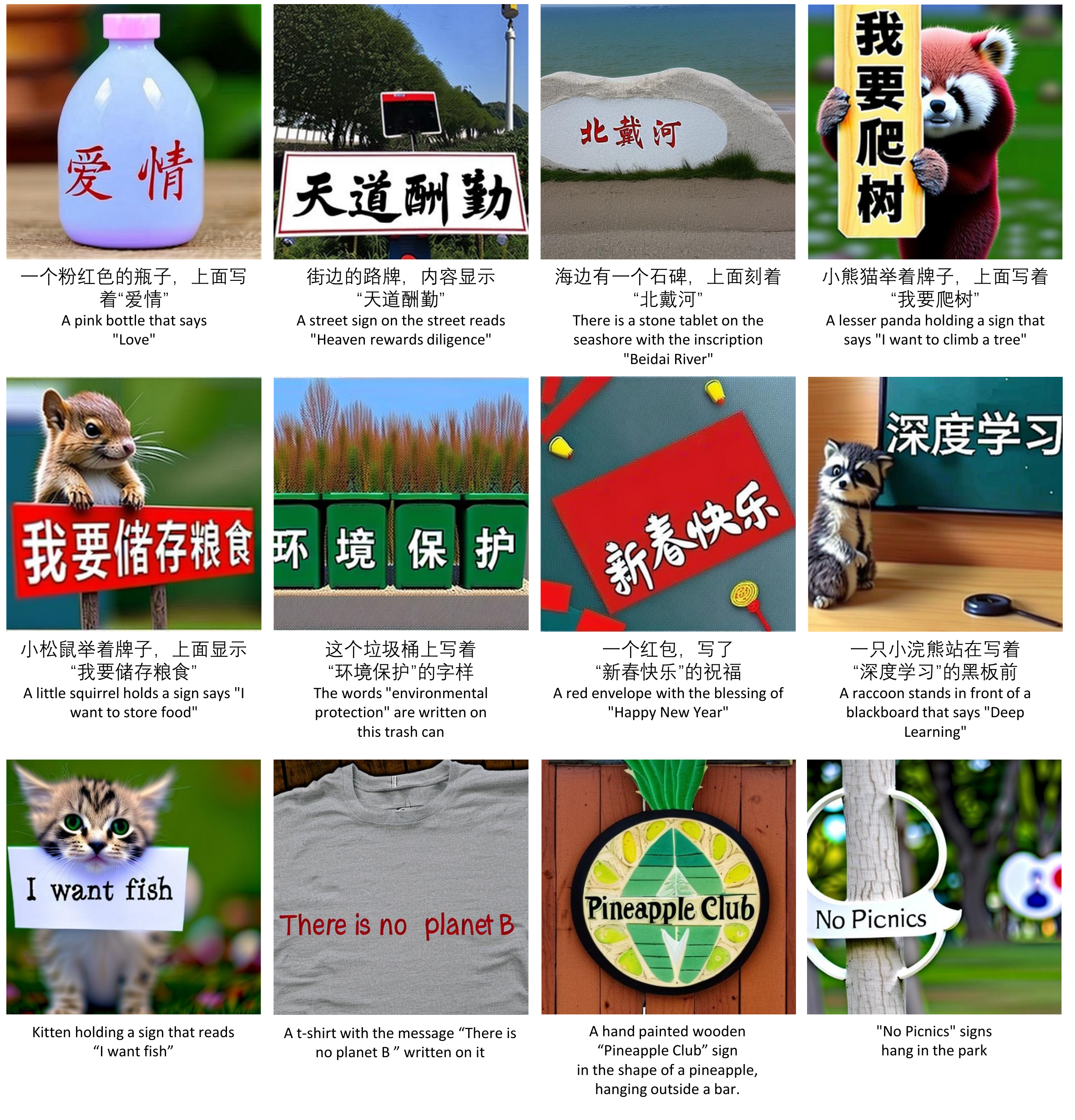}

\caption{We propose GlyphDraw, a general framework for endowing image generation models with the ability to generate high-quality visual text (\textit{e.g.} Chinese and English) embedded seamlessly in images while retaining their open-domain generation capability.}
\label{fig:demo}
\end{figure}

\begin{abstract}
Recent breakthroughs in the field of language-guided image generation have yielded impressive achievements, enabling the creation of high-quality and diverse images based on user instructions.
Although the synthesis performance is fascinating, one significant limitation of current image generation models is their insufficient ability to generate text coherently within images, particularly for complex glyph structures like Chinese characters. 
To address this problem, we introduce GlyphDraw, a general learning framework aiming to endow image generation models with the capacity to generate images coherently embedded with text for any specific language.
We first sophisticatedly design the image-text dataset's construction strategy, then build our model specifically on a diffusion-based image generator and carefully modify the network structure to allow the model to learn drawing language characters with the help of glyph and position information.
Furthermore, we maintain the model's open-domain image synthesis capability by preventing catastrophic forgetting by using parameter-efficient fine-tuning techniques.
Extensive qualitative and quantitative experiments demonstrate that our method not only produces accurate language characters as in prompts, but also seamlessly blends the generated text into the background.
Please refer to our \href{https://1073521013.github.io/glyph-draw.github.io/}{project page}.
\end{abstract}

\newpage

\section{Introduction}
The field of text-to-image synthesis has greatly benefited from recent advancements in diffusion models.
In contrast to generative adversarial network (GAN)-based methods \cite{goodfellow2020generative}, diffusion models offer greater stability during training without relying on the complex adversarial training process and enable precise control over generation quality and diversity during the diffusion process.
By leveraging the semantics of textual input, large-scale text-guided diffusion models such as DALLE-2 \cite{ramesh2022hierarchical}, Imagen \cite{saharia2022photorealistic}, and Stable Diffusion \cite{rombach2022high} have revolutionized the way we generate high-quality images.
These models are capable of following user instructions, generating customized high-fidelity images with specified content, style, and attributes. 

Recently, several studies \cite{zhang2023adding, mou2023t2i} have focused on gaining more fine-grained control over the generation process by leveraging auxiliary information such as edge maps, segmentation maps, color maps, \textit{etc.}
Users can specify desired attributes to guide and restrict the generation process by including such additional input, resulting in more precise control over the generated images.
Despite some limitations in image diversity due to auxiliary input dependency, these methods have produced impressive results in generating high-quality images that meet specific requirements.

Despite significant progress, current image synthesis methods face numerous challenges, particularly when it comes to generating fine-grained and intricate structures like human hands and textual content. A pioneer work Imagen \cite{saharia2022photorealistic} demonstrates that using frozen pretrained generic large language models (LLM), \textit{e.g.,} T5-XXL \cite{raffel2020exploring}, can render English text in images without introducing specifically-designed networks and training strategies. Recently, another work \cite{liu2022character} proposes to employ character-aware language models (ByT5 serials \cite{xue2022byt5}) to further enhance the visual text rendering ability of image synthesis models. However, as presented in \cite{liu2022character}, these methods are insufficient for generating non-Latin characters such as Chinese.
This is largely due to the distinctive characteristics of Chinese characters, which possess more complex two-dimensional spatial structures comprised of eight different types of basic strokes, as well as the large number of frequently used characters --- up to several thousands.
As a result, generating accurate and diverse Chinese characters poses a greater difficulty and remains an unresolved research problem. Furthermore, by freezing a pretrained LLM is inflexible to adapt a image synthesis model to render visual text of a user-specified down-stream language and training a specific LLM from scratch is prohibitively expensive and data-hungry. Therefore, we are inspired to design a general and adaptable algorithm to tackle the visual text rendering challenge by using light-weight training strategies and datasets.

To address this problem, we present GlyphDraw, a general framework designed to equip image generation models with the ability to generate coherent visual text within images.
GlyphDraw utilizes character glyphs and text locations in images as auxiliary information to provide greater control over the process of character generation. 
Our approach has yielded impressive results, with the ability to generate diverse visual text that precisely adheres to given instructions.
Notably, the generated text is intelligently matched with font style that best suits the context and integrated seamlessly into the background, while maintaining high generation quality and avoiding issues such as overfitting and catastrophic forgetting, as demonstrated by the Chinese and English examples in Fig. \ref{fig:demo}. Our main contributions are concluded as follows:
\begin{itemize}
\item We introduce GlyphDraw, a general and flexible framework to address the problem of visual character generation for any specified language (\textit{e.g.,} English or Chinese). 
GlyphDraw provides fine-grained guidance throughout the generation process, resulting in seamless integration of high-quality intricate characters with various styles into the image context.
\item We develop a parameter-efficient fine-tuning strategy based on the pre-trained model to prevent overfitting and catastrophic forgetting, thereby effectively preserving the model's strong open-domain generation performance while simultaneously enabling accurate visual text generation.
\item We delineate comprehensive training datasets' building procedure and evaluation benchmark, on which our GlyphDraw achieves excellent \textbf{74\%} and \textbf{75\%} OCR accuracy for Chinese and English characters rendering, respectively, significantly outperforming previous image synthesis approaches.
\end{itemize}

\section{Related Work}

\subsection{Text-to-Image Synthesis}
The diffusion model \cite{ho2020denoising,nichol2021improved,dhariwal2021diffusion,ho2022cascaded} has emerged as a promising direction to generate images with high fidelity and diversity based on provided textual input.
GLIDE \cite{nichol2022glide} introduces text condition into the diffusion process using classifier-free guidance.
DALL-E 2 \cite{ramesh2022hierarchical} adopts a diffusion prior module on CLIP text latents and cascaded diffusion decoder to generate high-resolution images. 
Imagen \cite{saharia2022photorealistic} emphasizes language understanding and proposes to use a large T5 language model for better semantics representation.
Latent diffusion model \cite{rombach2022high} projects the image into latent space with an autoencoder and applies the diffusion process to generate feature maps in the latent level.
Stable Diffusion \cite{rombach2022high} is an open-sourced model with excellent performance adopting the similar architecture.
We build our framework upon Stable Diffusion, although it can be easily transferred to other diffusion-based models with cross-attention mechanism as well.

\subsection{Condition-controlled Generation}
Other conditional information, in addition to text conditions, has been investigated to control image generation.
In-painting and out-painting techniques enable specific parts of an image to be modified based on user-provided masks and textual instructions while preserving the rest of the image \cite{avrahami2022blended,bau2021paint,nichol2022glide}.
In addition, Paint-by-example \cite{yang2022paint} proposes extracting the semantics from a reference image and using it to modify certain areas of an image.

Image-to-image translation deals with transforming images while preserving essential characteristics but changing other attributes such as style, color, and texture, and is dominated by conditional generative adversarial networks \cite{isola2017image,zhu2017toward,wang2018high,park2019semantic,choi2018stargan,zhang2020cross}.
Recently, diffusion-based image-to-image translation frameworks have gained much attention \cite{saharia2022palette,wang2022pretraining}.
ControlNet \cite{zhang2023adding} and T2IAdatper \cite{mou2023t2i} focus on providing a general solution to control the generation process using auxiliary information such as edge maps, color maps, segmentation maps, \textit{etc}, and deliver impressive results in terms of controllability and generated image quality. For our visual text rendering task, we are inspired to incorporate auxiliary glyph information into the conditional generation process and to emphasize local control throughout all diffusion steps.

\subsection{Draw Text in Image}
Numerous studies have explored the challenge of integrating textual content into image synthesis. For instance, studies on font generation aim to create novel fonts by treating it as a style translation problem based on a given input font \cite{cha2020few,jiang2019scfont,xie2021dg,zeng2021strokegan}.
Diff-font \cite{he2022diff} leverages diffusion models to handle the font generation task.
However, these works are only concerned with generating font glyphs without a background and do not correspond to our goal of improving text generation in image synthesis.
Another related work \cite{liu2022character} proposes character-aware diffusion models to improve text generation by incorporating character-level input features.
However, character-aware methods perform poorly for generating non-Latin text as mentioned in~\cite{liu2022character} due to 
the complexity of their spatial structures.
To the best of our knowledge, our paper presents the first work addressing the challenging problem of non-Latin (\textit{e.g.,} Chinese) character generation within general image synthesis.

\section{Methodology}
\label{sec:methodology}
In this section, we will first briefly review the necessary notations of Stable Diffusion \cite{rombach2022high} for a more convenient description of our proposed algorithm later on. Then, an overview of the GlyphDraw framework followed by an explanation of how we leverage auxiliary information will be detailed\footnote{For brevity of description, without loss of generality, we adopt Chinese as example throughout Sec. \ref{sec:methodology}.}. Further more, we will introduce our training strategies designed to prevent catastrophic forgetting. Finally, we will present the inference process, which differs slightly from the training stage.

\subsection{Preliminaries}
In this paper, we implement our method based on an open-source image synthesis algorithm -- Stable Diffusion \cite{rombach2022high}\footnote{It's worth noting that our approach is a general framework which can be easily transplanted into other diffusion-based methods by employing cross-attention mechanism, such as ERNIE-ViLG 2.0~\cite{feng2022ernie} and eDiff-I~\cite{balaji2022ediffi}.}, which enhances computational efficiency by performing the diffusion process in low-dimensional latent space with the help of an auto-encoder. 
Specifically, for an input image $x_0 \in \mathbb{R}^{H \times W \times 3}$, the encoder $\mathcal{E}$ of the auto-encoder transforms it into a latent representation $z_0 \in \mathbb{R}^{h \times w \times c}$, where $f=H/h=W/w$ is the downsampling factor and $c$ is the latent feature dimension.
The diffusion process is then performed on the latent space, where a conditional UNet~\cite{ronneberger2015u} denoiser $\epsilon_\theta$ is employed to predict noise $\epsilon$ with current timestep $t$, noisy latent $z_t$ and generation condition $C$.
The condition information $C$ is fed into each cross attention block $i$ of the UNet model as
\begin{equation}
\label{eq:sd}
\begin{aligned}
    &{Attention}(Q, K, V) = {softmax}\left(\frac{QK^T}{\sqrt{d}}\right)\cdot V \\
    \textit{where} \quad &Q = W_Q^{(i)} \cdot \varphi_i(z_t), K = W_K^{(i)} \cdot C, V = W_V^{(i)} \cdot C.
\end{aligned}
\end{equation}
Here, $d$ denotes the output dimension of key ($K$) and query ($Q$) features, $\varphi_i(z_t)$ is a flattened intermediate representation of the noisy latent $z_t$ through the UNet implementation $\epsilon_\theta$, and $W_Q^{(i)}, W_K^{(i)}, W_V^{(i)}$ are learnable projection matrices. 
In text-to-image scenarios, the condition $C=\tau_\theta(y)$ is produced by encoding the text prompts $y$ with a pre-trained CLIP~\cite{radford2021learning} text encoder $\tau_\theta$. Therefore, 
the overall training objective of Stable Diffusion is defined as
\begin{equation}
    \mathcal{L}_\text{SD} = \mathbb{E}_{\mathcal{E}(x_0),C,\epsilon \sim \mathcal{N}(0,1),t} \Big[ \lVert \epsilon - \epsilon_\theta (z_t, t, C)\rVert_2^2 \Big]
\end{equation}

\subsection{Model Overview}
\begin{figure}[tb]
	\centering
	\includegraphics[width=.9\textwidth]{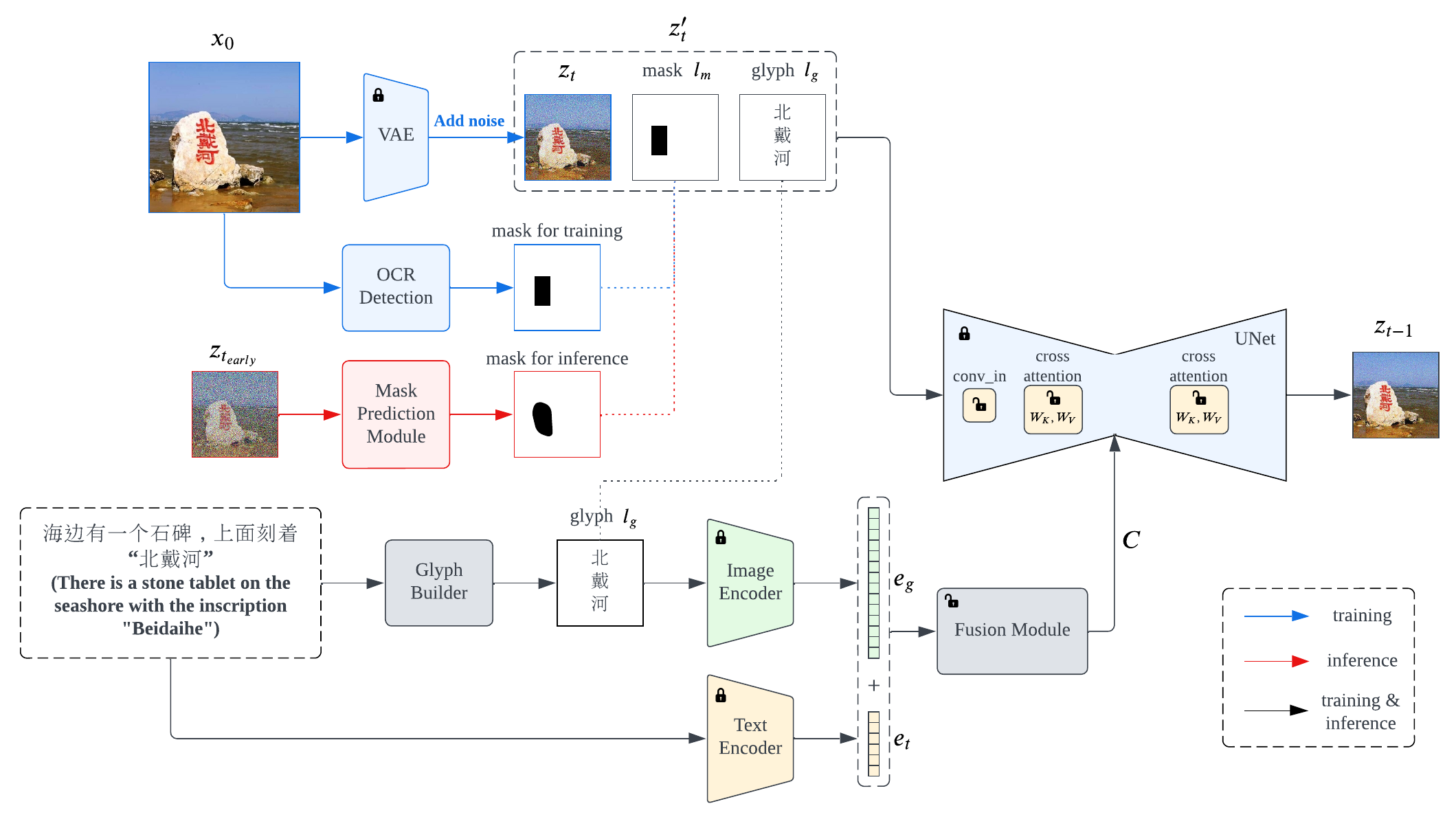}
\caption{An overview of the proposed GlyphDraw method based on Stable Diffusion structure. For the image latent part, an OCR detected character mask $l_m$ and a glyph image $l_g$ containing only character visual information are concatenated to image latent feature $z_t$. Then the combined latent feature $z'_t$ is used as input of the UNet. As to the text condition part, a pretrained CLIP model encodes the prompt and glyph image into embeddings $e_t$ and $e_g$, respectively. Then, a fusion module is employed to further fuse the text and glyph embeddings to a conditional feature $C$ which is used as the key and value components of UNet cross-attention layers. During inference, an MLP-like mask prediction module is adopted to estimate the character mask map.}
\label{fig:Framework}
\end{figure}

The overall training framework of our proposed GlyphDraw method is illustrated in Fig.~\ref{fig:Framework}. We focus on modifying the cross attention mechanism in Stable Diffusion. The original input latent vector $z_t$ is replaced by the concatenation of image latent vector $z_t$, text mask $l_m$ and glyph image $l_g$. In addition, the condition $C$ is equipped with blended glyph and text feature by employing a domain specific fusion module. The introduction of text mask and glyph information, which enable fine-grained diffusion control throughout the training process, is one of the key performance-enhancing components.



\subsection{Exploitation of Auxiliary Information}
The pixel representation of textual information, especially ideograph-like Chinese character, differes significantly from that of natural objects. For instance, the Chinese word ``\begin{CJK}{UTF8}{gbsn}天空\end{CJK}(sky)'' is simply composed of multiple strokes in a two-dimensional structure, whereas the imagination of its natural image counterpart is like ``\textit{a giant blue screen dotted with white clouds}''. Visual text is a kind of very fine-grained feature and even a minor shift or deformation will result in incorrect textual rendering, leading to unrealistic image generation. Another critical issue to consider when embedding characters in natural image backgrounds is how to precisely control textual pixel generation while avoiding affecting neighboring natural image pixels. Therefore, in order to render perfect characters on realistic natural images without incongruity, we meticulously design two pivotal components integrated into diffusion-based synthesis model, namely \textit{location control} and \textit{glyph control}.

\textbf{Location Control}.
Unlike global conditional input such as segmentation maps, depth maps, sketches and grayscale images as in~\cite{zhang2023adding}, character generation requires more attention to a specific local area of the image, since the latent feature distribution of character pixels varies dramatically from that of natural image pixels. To prevent model learning from collapsing, a fine-grained location area control is innovatively introduced to decouple the distribution between different regions. Specifically, as shown in Fig.~\ref{fig:Framework}, a binary mask feature map is generated and concatenated to original image latent feature. During the training phase, the quadrilateral-shaped mask is extracted from OCR detection \cite{du2021pp} information, which will be detailed described in Appendix~\ref{app:dataset_details}. During inference phase, since there is no available reference image, the mask is generated through a mask prediction module during early diffusion stage, which will be further discussed in Sec.~\ref{method:inference}.


\setcounter{footnote}{0} 
\textbf{Glyph Control}. 
Apart from location control, as mentioned above, another important challenge is the fine control of language character stroke synthesis. Given the complexity (typically composed of several 2D strokes ranging from 1 to 20) and variety (up to 10,000 common characters) of Chinese characters, naively learning from massive image-text datasets without any explicit prior knowledge injection is extremely difficult. As demonstrated in~\cite{liu2022character}, character-blinded model can induce robust spelling knowledge for English words only when model parameters are greater than 100B, and even character-aware model cannot generalize well beyond Latin scripts such as Chinese, Japanese and Korean. In addition, we also directly fine-tune a Chinese Stable Diffusion checkpoint (more details in Appendix~\ref{app:sd_checkpoints}) on our proposed textual image dataset but achieve completely fail results as presented in Sec.~\ref{exp:comparison}.
Therefore, to generate Chinese characters accurately, we are inspired to incorporate explicit glyph images as extra condition information into the model diffusion process. Specifically, as illustrated in Fig.~\ref{fig:Framework}, a pre-extracted glyph image containing only Chinese characters located in the image center with a white background, \textit{e.g.,} ``\begin{CJK}{UTF8}{gbsn}北戴河\end{CJK}(Beidai River)'', is simultaneously injected into both the image latent (above flow) and text embedding (below flow) parts. 

First, the grayscale glyph image $l_g$ extracted by the glyph builder is concatenated to the noisy image latent vector $z_t$ and the binary text mask $l_m$ to form a new latent vector $z'_t=concat(z_t, l_g, l_m)$. After dimension adjustment through a convolution layer, the feature vector $\Tilde{z}_t=conv\_in(z'_t)$ is fed into the UNet as the query component. In terms of the conditional information, $C=\mathbb{M}\left[concat(e_g, e_t)\right]$ is fused by a fusion module $\mathbb{M}$ from glyph embedding ($e_g = I_\theta(l_g)$) and text embedding ($e_t = \tau_{\theta}(y)$) which are extracted from fixed CLIP image encoder ($I_\theta$) and text encoder ($\tau_{\theta}$), respectively. Consequently, the basic GlyphDraw training objective is:
\begin{equation}
\label{eq:loss}
\mathcal{L}_\text{GD\_b} = \mathbb{E}_{\mathcal{E}(x_0),y,l_g,l_m,\epsilon \sim \mathcal{N}(0,1),t} \big[\parallel\epsilon-\epsilon_\theta(z_t,t,y,l_g,l_m)\parallel^2_2\big].
\end{equation}

\subsection{Training Strategies}
\begin{wrapfigure}{1}{.5\textwidth}
    \centering
	\includegraphics[width=0.5\textwidth]{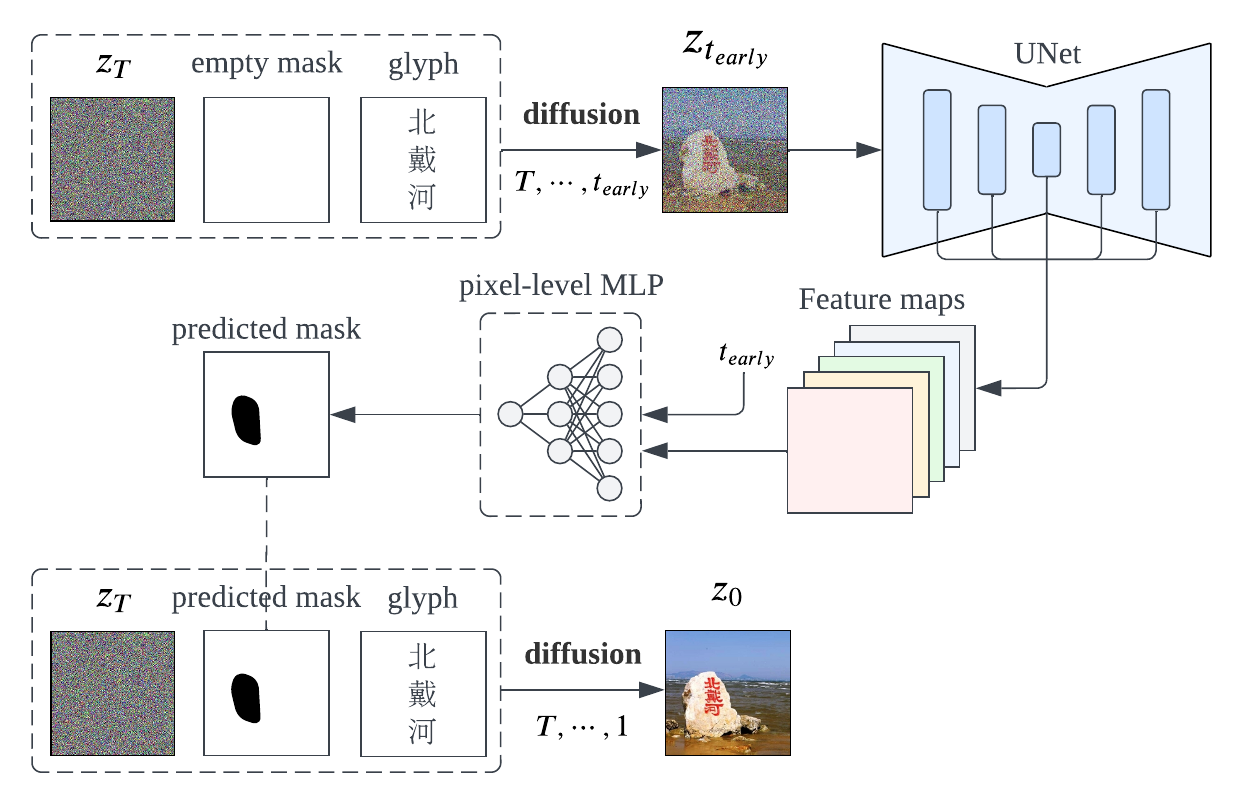}
	\caption{The design of the mask prediction module. During inference process, we initialize empty mask for the first few diffusion steps ($t=\{T,T-1,\dots,t_{early}\}$) and employ a pixel-wise MLP to estimate the mask. Then we adopt the predicted mask to generate the final desired images through a new complete diffusion process ($t=\{T, T-1, \dots, 1\}$).}
	\label{fig:MaskPred}
\end{wrapfigure}
\textbf{Anti-catastrophic forgetting}.
Fine-tuning an image generation model can often lead to overfitting and catastrophic forgetting, which significantly degrades the model's open-domain performance.
To address these issues, we develope a parameter-efficient fine-tuning strategy that preserves the original model's ability to the greatest extent possible.
In other words, during the learning stage, we only update the network parameters required for learning language character generation while freezing other parameters to preserve the model's general ability.

To enable the UNet to take the location mask and glyph information as two additional channels alongside the image latent, we adjust the input ``conv\_in'' module of the UNet accordingly to accommodate the additional information and update it during learning. 
Similarly, the fusion module, which modifies the generation condition $C$ by integrating the embeddings of glyph information and text prompts, also needs to be updated.
Moreover and most importantly, as proposed in~\cite{radford2021learning}, when updating the mapping from given text to image distribution, only updating $W_K^{(i)}, W_V^{(i)}$
in each cross-attention block $i$ is sufficient since text features are the only input to key and value projection matrix.
By carefully choosing the parameters to update, our method effectively maintains the model's generation performance and enables coherent text generation while updating only \textbf{3\%} of the total parameters, significantly accelerating model convergence.

\textbf{Loss weighting}.
To further enhance the visual text generation performance of our model, we implement a weighting strategy in the design of the training objective, with the goal of emphasizing the learning of language character generation ability during the learning process.
Specifically, we add extra weight to the training loss of the text area based on the location mask information $l_m$, as follow:
\begin{equation}
\label{eq:weighting_loss}
\begin{aligned}
    \mathcal{L}_{GD} \!=\! \mathbb{E}_{\mathcal{E}(x_0),y,l_g,l_m,\epsilon \sim \mathcal{N}(0,1),t} \Big[ \lVert \epsilon \!-\! \epsilon_\theta (z_t, t, y, l_g, l_m)\rVert_2^2 \!+\! \alpha \lVert \big[\epsilon \!-\! \epsilon_\theta (z_t, t, y, l_g, l_m) \big] \!*\! (1\!-\!l_m) \rVert_2^2 \Big], \\
\end{aligned}
\end{equation}
where $\alpha \ge 0$ is a weighting hyper-parameter.

\subsection{Inference}
\label{method:inference}
During inference, the mask information $l_m$ cannot be directly extracted by OCR detectors since the absence of original image $x_0$. Thus, we propose a mask prediction module in Fig.~\ref{fig:Framework} (red lines and boxes) to estimate a rough mask with arbitrary shape. As elaborately depicted in Fig.~\ref{fig:MaskPred},  we estimate the character mask during first few diffusion steps ($t=\{T, T-1, \dots, t_{early}\}$) by using a simple pixel-wise MLP network trained by MSE loss between estimated and ground-truth masks. After obtaining the predicted mask, we regenerate images through a complete diffusion process ($t=\{T, T-1, \dots, rT, \dots, 1\}$) by DDIM~\cite{song2021denoising} sampling strategy. We sample the first few steps ($\{T, T-1, \dots, rT+1\}$) by our Glyphdraw model and the remaining steps ($\{rT,\dots, 1\}$) by the pretreained Stable Diffusion model, discarding glyph and location priors, where $r\in [0, 1]$ is a hyper-parameter to trade text rendering accuracy for open-domain generation capability. 

If no textual information is required to be rendered in the image, the mask is simply set to 1, resulting in a completely white mask.
Furthermore, we can accept user-supplied location masks as well as randomly generated masks. But compared to the random mask, the estimated masks after training satisfy the perspective theory \cite{szeliski2022computer} in projective geometry, making the rendered visual text look more realistic from the viewing angle. More ablation and case studies will be discussed in Sec. \ref{sec:ablation}.

\section{Experiments}

\subsection{Data Preparation}
Due to a lack of existing dataset for image synthesis with visual text generation, we gather data from various sources, including web crawled data and publicly available image-text datasets including LAION \cite{schuhmann2022laion}, Zero23M \cite{xie2022zero}, and Wukong \cite{gu2022wukong}, to prepare our training dataset.

In order to collect high-quality images embedded with visual text, we adopt a data pre-processing procedure to filter data and extract glyph and location information.
Specifically, we employ an OCR model \cite{du2021pp} to locate and recognize language characters (\textit{e.g.} English or Chinese) in images, and apply a data filtering criterion to reduce the potential noise in datasets.
In addition, we also discover that the quality of captions varies drastically.
Thus, we exploit the BLIP-2 model \cite{li2023blip} to regenerate captions for the collected images and embed the characters by quotation marks into the captions.
More data filtering details, visualization examples and statistics of the cleaned datasets are presented in Appendix~\ref{app:dataset_details}. After painstaking filtering, we finally collect 792k images for the Chinese dataset, containing 3.3M characters in images and more than 4.8k common unique Chinese characters; for the English dataset, 1.9M images are gleaned and 2.3M English words are extracted.

\subsection{Implementation Details and Evaluation}
\label{implem_details_eval}
\textbf{Implementation details}.
Based on Stable Diffusion, GlyphDraw consists of VAE, UNet, CLIP, and the fusion module, comprising 1.9 billion parameters, out of which a mere 0.1 billion parameters (fusion module, \textit{conv\_in} module and projection matrices $W_K^{(i)}, W_V^{(i)}$) are trainable. The VAE and UNet are initialized from Stable Diffusion checkpoints 
and the CLIP image and text encoder are loaded from pretrained CLIP checkpoints (more details of the pretrained checkpoints are presented in Appendix \ref{app:pretrained_checkpoints}).
After CLIP encoders, the image and text token lengths are 257 and 64, respectively. We adopt a transformer with 6 layers, 8 attention heads and 1024 hidden dimension as the fusion module. We set the learning rate to 2e-5 and the weighting scale hyper-parameter $\alpha$ in Eq.~(\ref{eq:weighting_loss}) to 0.5. The entire model is trained on 24 A100 GPUs for 20 epochs with batch size of 25 per GPU.

During inference stage, we adopt a 5-layer MLP for the mask prediction module, and set $T=50$,  $T_{early}=35$ and $r=0.5$ in Sec. \ref{method:inference} (using first 15 steps to generate the estimated mask; sample first 25 steps by GlyphDraw and another 25 steps by Stable Diffusion to generate realistic images). We also illustrate the OCR accuracy and FID comparisons over different $r$ as in Appendix \ref{app:acc_fid_tradeoff}.

\textbf{Test benchmark}. 
Following the benchmark \textit{DrawText} proposed in \cite{liu2022character}, we extend it to support both English and Chinese, called \textit{DrawTextExt}. Similarly, \textit{DrawTextExt} consists of \textit{DrawTextExt Spelling} and \textit{DrawTextExt Creative} for quantitative and qualitative, respectively.

For DrawTextExt Spelling, a prompt with standard template is divided into two parts: a fixed prefix sentence ``\begin{CJK}{UTF8}{gbsn}街边的路牌上写着*\end{CJK} (The sign on the street says *)'' is combined with the characters to be embedded, such as ``\begin{CJK}{UTF8}{gbsn}天道酬勤\end{CJK} (heaven rewards diligence)''. We further substitute the character description in prompts by other words, \textit{e.g.,} replace ``\begin{CJK}{UTF8}{gbsn}天道酬勤\end{CJK} (heaven rewards diligence)'' by ``\begin{CJK}{UTF8}{gbsn}我要学习\end{CJK} (I want to study)'', to expand total dataset size, resulting in 10,000 common Chinese characters (including words) \cite{chinese-common-vocabulary} and 10,000 high-frequent English words \cite{high-frequency-vocabulary}.
For DrawTextExt Creative, the dataset is comprised of 200+ various creative prompts collected in two ways: extracted and translated from a subset of the DrawText dataset and manually written by ourselves, which can stimulate both Chinese and English natural scenes to the greatest extent possible. All the prompts are presented in Appendix~\ref{app:test_prompts} for reference. 
 
\textbf{Evaluation metrics}.
We utilize a well-performing OCR tool \cite{du2021pp} to calculate the character recognition accuracy on above test benchmarks.
However, accuracy alone does not faithfully represent the quality of the generated images, which may result in overfitting or forgetting and thus poor image generation quality. Therefore, we also employ the widely used MS-COCO FID-10k~\cite{heusel2017gans} metric to evaluate the overall image generation quality.

\textbf{Baseline methods.}
Since only limited works concern about the visual text rendering task, we compare our algorithm to two intuitive methods and Imagen \cite{saharia2022photorealistic} as described below.

\begin{itemize}
    \item Fine-tuning.
A straightforward approach is to fine-tune a pre-trained image generation model (Stable Diffusion) on a specific text-image dataset and evaluate its ability to generate desirable visual text. 

    \item ControlNetDraw.
ControlNet \cite{zhang2023adding} is an image synthesis method that achieves excellent controllability by incorporating additional conditions to guide the diffusion process. 
Inspired by its great control over image generation, we adapt this method to text rendering problem by taking the glyph image as the input to the ControlNet network. We call this method ControlNetDraw.

    \item Imagen \cite{saharia2022photorealistic}. 
As mentioned in \cite{liu2022character}, Imagen can infer robust spelling information through pretrained LLM knowledge (T5 serials). We employ a recently released DeepFloyd IF \cite{deep-floyd-IF} open source for English text rendering comparison.

\end{itemize}



\subsection{Comparison results}
\label{exp:comparison}
Table \ref{tab:experiment_stats} compares the OCR accuracy and FID of various methods on our proposed DrawTextExt dataset, revealing that stable diffusion fails to generate correct characters, resulting in an accuracy of almost 0\%. Even vanilla fine-tuning and the modified ControNetDraw struggle to enhance OCR accuracy, which are completely ineffective for the Chinese model and achieve only marginal improvements (15\% and 7.2\%) for the English model.
Visualization results in Appendix \ref{app:visualization_comp} indicate that these two methods cannot capture the precise spatial structure of the characters and produce incorrect strokes.
Their failures can be attributed to two reasons: 1) using direct fine-tuning to generate characters based on the semantic representation of the prompts without access to the glyph information of the characters is extremely challenging; 2) ControlNet aims to control the image generation with an exact mapping from the input condition to the output image, such as edge maps. However, in the text rendering task, there is an inevitable gap between the provided glyph and the text in target images due to the variations in font, size and perspective, which greatly degrades the performance of ControlNet. Imagen demonstrates that LLM indeed enhances the implicit character-level understanding capability of the image generation model, but is still inferior to our explicit glyph and location control strategy, resulting in only 44.5\% accuracy for the English model.


    \label{tab:fid_results}

\begin{table}[tbp]
\renewcommand{\arraystretch}{0.9}
\centering
\caption{OCR accuracy and FID values for all comparisons and ablation studies, including component and training parameter ablations. We run OCR evaluations for three models with different random seeds and report the averages and standard deviations.}
\resizebox{1.0\columnwidth}{!}{
\begin{tabular}{c|ccccc}
\toprule
& & \multicolumn{2}{c}{Chinese Model} & \multicolumn{2}{c}{English Model} \\ 
\cmidrule(l){3-6} 
Category & Method & Accuracy ($\uparrow$ \%) & FID ($\downarrow$)  & Accuracy($\uparrow$ \%) & FID ($\downarrow$)  \\
\midrule
& Stable Diffusion \cite{rombach2022high} & 0.00$\pm$0.00 & 15.87 & 0.14  & 14.69   \\
baselines & Stable Diffusion + Fine-tuning & 0.00$\pm$0.00 & 20.09 & 15.01$\pm$0.64 & 20.24 \\
& ControlNetDraw & 0.00$\pm$0.00 & 15.91 & 7.18$\pm$0.41 & 14.63\\
& Imagen \cite{saharia2022photorealistic} & - & - & 44.54$\pm$0.68 & - \\
\midrule
& Glyphdraw w/o loss weighting & 68.60$\pm$1.09 & 17.27 & 64.96$\pm$0.32 & 16.92 \\
component & Glyphdraw w/o location mask & 66.15$\pm$0.89  & 17.68  & 65.79$\pm$0.70  &17.85 \\
ablations & Glyphdraw w/ random mask & 60.23$\pm$0.18 & 16.78 & 60.55$\pm$1.06 & 16.09 \\
& Glyphdraw w/o glyph latent & 69.88$\pm$0.52 & 17.36 & 65.21$\pm$0.57 & 16.58 \\
\midrule
& Glyphdraw + all UNet layers & 74.21$\pm$0.83 & 19.01 & 77.29$\pm$0.82 &19.78\\
training parameter & Glyphdraw + all UNet attention layers & 72.24$\pm$0.74 & 17.95 & 76.74$\pm$0.83 & 18.01\\
ablations & Glyphdraw (only $W_k^{(i)}$ and $W_v^{(i)}$) & 74.00$\pm$1.13 & 16.89 & 75.23$\pm$1.06 & 16.02\\
\bottomrule
\end{tabular}\label{tab:experiment_stats}
}
\end{table}

In contrast, our proposed GlyphDraw achieves 74\% and 75\% average accuracy over the same quantitative testset for Chinese and English, respectively, by effectively using the auxiliary glyph and location information, demonstrating its outstanding character generation ability. 
Additionally, GlyphDraw can also maintain open-domain image synthesis performance by constraining the training parameters, achieving only a slight FID degradation on general image synthesis measured on MS-COCO FID-10k (1.02 for Chinese and 1.33 for English). Appendix \ref{app:images_no_text} shows some general images containing no visual text generated by GlyphDraw.

\subsection{Ablation studies}
\label{sec:ablation}
The Ablation studies involve examining three main aspects, namely 1) the impact of local control and glyph control on the image generation performance of GlyphDraw, 2) the effectiveness of training strategy for mitigating catastrophic forgetting, and 3) the inference performance affection by using mask prediction module. All statistics of OCR accuracy and FID value is provided in Table~\ref{tab:experiment_stats}.

\textbf{Loss weighting}.
The loss weighting strategy enables the model to focus more on the generation quality of the text area by assigning higher weights on the loss of corresponding area, which generally improves the text generation accuracy by a large margin (5.4\% for Chinese and 11.3\% for English) and also slightly optimizes FID compared to the model variant without loss weighting.

\textbf{Location mask}.
When eliminating location mask, only glyph image is concatenated to the input noisy image latent vector. Without such explicit local guidance, The OCR accuracy significantly reduces (more than 8\% drop for both Chinese and English) with respect to GlyphDraw, especially for scenario that characters number is smaller than 3, possessing only a small portion of the image (detailed diagrams of accuracy over character length are provided in Appendix \ref{app:acc_detail}).

\textbf{Mask prediction module}.
The mask prediction module aims to provide an estimated mask of arbitrary shape for text generation during the inference stage. 
There are two advantages for incorporating the mask prediction module.
First, compared to using a random mask, the learned module can better capture the shapes of the text areas and guide the text generation more effectively, leading to a remarkable 15\% higher OCR accuracy as shown in Table \ref{tab:experiment_stats}.
Second, the module is able to effectively learn the transformation effect due to different viewing perspectives and produces images that are more realistic and natural (see the visualization examples  in Appendix \ref{app:visualization_comp}).

\textbf{Glyph latent}.
In the design of GlyphDraw, the glyph information is simultaneously used in the latent $z_t'$ as $l_g$ and the conditional information $C$ as $e_g$. 
To assess significance of this design, we conducted an evaluation by removing $l_g$ while retaining $e_g$ in the model.
The results indicate an apparent OCR accuracy decrease (4\% for Chinese and 10\% for English), thus confirming the effectiveness of incorporating glyph information at multiple levels in enhancing model's character-level understanding.

\textbf{Ablations on continual training strategies}. 
We compare different continual training strategies --- fine-tuning the entire UNet network, with 865M parameters, fine-tuning all UNet attention layers  with 100M parameter, accounting for 12\% of total UNet parameters and fine-tuning only the cross-attention key and value projection matrices of the UNet with 25M parameter, accounting for only 3\% of total UNet parameters. Final results in Table~\ref{tab:experiment_stats} reveals that fine-tuning only K and V cross-attention components achieves minimal parameter modification, best FID performance, and ignorable OCR accuracy degradation (0.2\% for Chinese and 2\% English compared to fine-tuning all UNet parameters), leading to the strongest anti-forgetting capability and the fastest convergence speed.

\section{Conclusion and Limitation}
In this paper, we present GlyphDraw, a visual text generation framework designed to address the limitations of existing image synthesis models in producing high-quaity and coherent text embedded in natural images. 
By leveraging glyph and location information, GlyphDraw offers fine-grained control over the generation process, allowing for the accurate generation of desired characters according to user instructions. 
Moreover, the framework intelligently blends the generated text into image background seamlessly, maintaining high generation quality and natural congruity. It can adapt to any user-specified language such as Chinese and English based on provided training dataset.
Overall, through extensive experiments, we demonstrate that GlyphDraw has the potential to deliver remarkable performance, both in terms of objective OCR accuracy and subjective visual quality.

Our method is subject to a limitation that the generation accuracy decreases as the number of characters to be drawn in the image increases. 
This is due to the fact that as the number of characters increases, so does the spatial complexity, making the generation process more challenging. 
Furthermore, our training dataset consists primarily of images with short visual text, with an average character count of 4.17 and letter count of 7.19 for Chinese and English, respectively. 
As a result, the model's ability to generate long text in images is restricted. Another limitation of our proposed method is that it does not support multiple text bounding-box generation. Consequently, our future work will focus on enhancing generation quality, diversity, tackling the long-text rendering issue and synthesizing more creative and aesthetic images.

\bibliographystyle{ieee_fullname}
\bibliography{main}


\newpage
\section*{Appendix}
\appendix
In this supplementary, we will first introduce the details of pretrained checkpoints employed in our GlyphDraw approach in Appendix \ref{app:pretrained_checkpoints}, including Chinese and English stable diffusion and CLIP models. In Appendix \ref{app:dataset_details}, we will present more details of the training datasets, containing data collection sources, data filtering strategies, image caption regeneration and dataset statistics for both Chinese and English. Then in Appendix \ref{app:visualization}, we provide more subjective visualization examples for visual text rendering results, synthesized images without text and ablations and comparisons with other methods, demonstrating GlyphDraw's visual text generation capability and open-domain generalization. Additionally, we also depict the trade-off between OCR accuracy and FID over hyper-parameter $r$ mentioned in Sec. \ref{method:inference}. Finally, Appendix \ref{app:test_benchmark} present the OCR evaluation criteria, detailed results, and the DrawTextExt test prompts for reference.

\section{Pretrained checkpoints}
\label{app:pretrained_checkpoints}
\subsection{Stable diffusion checkpoints}
\label{app:sd_checkpoints}
We pretrain our Chinese stable diffusion model based on the Stable Diffusion v2-base model checkpoint\footnote{\url{https://huggingface.co/stabilityai/stable-diffusion-2-base}} by only fine-tuing the text encoder. The training dataset is made up of the Chinese part of Laion-5B, the translated version of Laion-5B, Noah-Wukong, and some self-crawled image-text pairs. We create a 100M Chinese image-text dataset after strict filtering conditions, such as removing images and prompts with low quality, images with low aesthetic score or image resolution less than 512$\times$512. Finally, we train the text encoder with 160K steps and a batch size of 2000 on 80 A100 GPUs.

In terms of English stable diffusion, we directly employ the Stable Diffusion v2-base model checkpoint.

\subsection{CLIP checkpoints}
\label{app:CLIP_checkpoints}
For the Chinese checkpoint, we fine-tune the Chinese CLIP\footnote{\url{https://github.com/OFA-Sys/Chinese-CLIP}} on our collected dataset as our text encoder. While for the English experiment, we directly used the original Stable Diffusion pretrained text encoder (OpenCLIP-ViT/H\footnote{\url{https://huggingface.co/laion/CLIP-ViT-H-14-laion2B-s32B-b79K}}).

\section{Dataset preparation details}
\label{app:dataset_details}
In order to prepare the training dataset, we have applied several procedures to ensure the scale and the quality of the collected dataset, including data collection, data filtering and caption generation.

\subsection{Data collection.}
To generate text in images, we need to collect a large number of images containing visual text for the model learning.

For English datasets, we incorporate the 600M high-quality images with predicted aesthetic scores greater than or equal to 5 from LAION-Aesthetics V2\footnote{\url{https://laion.ai/blog/laion-aesthetics/}}.

For Chinese datasets, we adopt large-scale multi-modality Chinese datasets Noah Wukong~\cite{gu2022wukong} and Zero23M~\cite{xie2022zero} containing 100 million and 23 million samples respectively, as well as the Chinese part in LAION~\cite{schuhmann2022laion} which consists of 143 million samples.
Apart from these samples, we also crawled 86 million images from the web using search engines to enrich the dataset. 

All datasets will be publicly released upon paper acceptance.

\subsection{Data filtering.}
Following data collection, we implement a comprehensive data filtering procedure to extract high-quality images containing text. We utilize an OCR model \cite{du2021pp} to detect and recognize text in the images and have designed a set of rules to filter out low-quality samples. These rules are outlined as follows:

\textbf{Limitation on recognition confidence.}
To minimize the impact of OCR model misrecognition, we have established a threshold value of 0.8 on the confidence score of the OCR recognition results. Recognition results with lower confidence are more likely to be incorrect and are therefore discarded.

\textbf{Constraints on text size and location.}
For ease of learning, we want the text to be a significant part of the images, that is said the text should not be too small or locate at an easily overlooked corner. 
Therefore, we set a limit on the size of each Chinese/English character in the text to be at least 0.7\%/0.2\% of the image size, and the center location of the text to be away from the edges by at least 10\% of the width/height of the image.
We have also filtered out the images with multiple text, which could confuse the model learning on where to generate the text.

\textbf{Falsely identified characters.}
Some characters have special spatial structures that often lead to false identification, for example, the character \begin{CJK}{UTF8}{gbsn}"田"\end{CJK} is frequently misidentified in images containing windows or grids. To address this issue, we have compiled a list of easily misidentified characters and discard images that only contain characters from this list.
The list for Chinese characters is shown as: \begin{CJK}{UTF8}{gbsn}["米", "口", "回", "人", "王", "川", "大", "美", "三", "丰", "区", "中", "十", "田", "山", "一", "下", "个", "门", "八", "小", "品", "具", "工"]\end{CJK}.
For English, we discard images with only one letters.

\textbf{Advertisement filtering.}
We also found that many images with Chinese text are ads for items on sale, where the image quality is low and the text is artificially generated and poorly pasted on the image. To filter out these images, we made another keyword list with common e-commerce terms: \begin{CJK}{UTF8}{gbsn}["厂价", "直销", "包邮", "包赔", "立减", "清仓", "买1", "买一", "已售", "客服", "拍下", "改价", "开票", "厂家", "质保", "超值", "礼包", "限时", "全赔", "系列", "新品"]\end{CJK}.

We also got rid of images with website link watermarks as they are often considered as noise in visual text generation.

\subsection{Caption Generation}
We regenerated the captions for derived images for two reasons. First, webcrawled images do not have captions. Second, the original captions in these datasets vary in quality and often do not match the image content. Therefore, we used the state-of-the-art image captioning model BLIP-2~\cite{li2023blip} to generate new captions.
For Chinese data, we translated generated captions into Chinese with a translation system.

\subsection{Dataset statistics}

\begin{table}[htb!]
    \centering
    \subfloat[Chinese training data statistics.]{
    \begin{tabular}{c|ccc}
        \toprule
        Filtered Dataset & \# Samples & \# Chars & \# Unique Chars \\
        \midrule
        Laion & 130,024 & 554,500 & 4,236 \\
        Zero23M & 101,528 & 430,839 & 4,088 \\
        Noah-Wukong & 262,063 & 1,111,885 & 4,487 \\
        Web crawled & 298,689 & 1,203,591 & 4,414 \\
        \midrule
        Total & 792,304 & 3,300,815 & 4,816 \\
        \bottomrule
    \end{tabular}
    }
    \quad
    \subfloat[English training data statistics.]{
    \begin{tabular}{c|cccc}
        \toprule
        Filtered Dataset & \# Samples & \# Chars & \# Words & \# Unique Words \\
        \midrule
        LAION-Aesthetics V2 & 1,929,981 & 13,870,183 & 2,339,693 & 339,055 \\
        \bottomrule
    \end{tabular}
    }
    \quad
    \subfloat[An example of our Chinese training data.]{
    \begin{tabularx}{\textwidth}{ccc>{\centering\arraybackslash}X}
        \toprule
        Image & Glyph & Mask & Caption \\
        \midrule
        \adjustbox{valign=t}{\frame{\includegraphics[width=.2\textwidth]{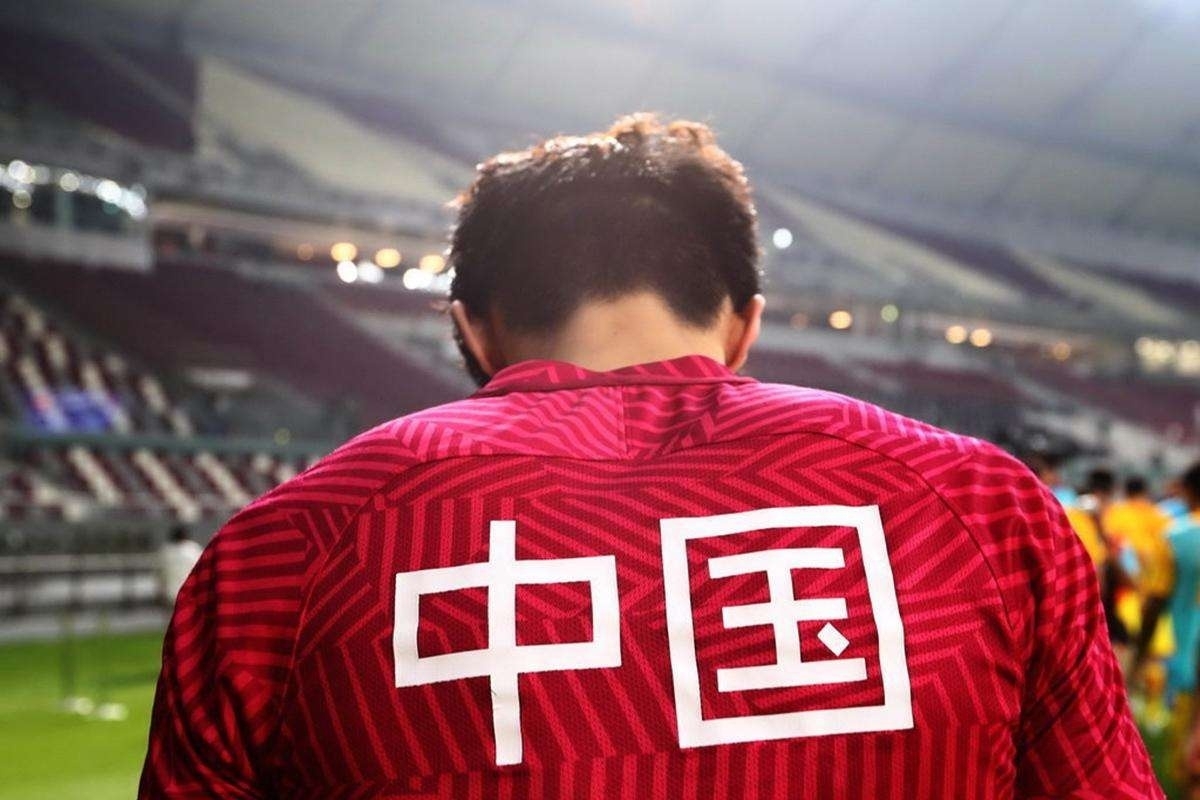}}} & 
        \adjustbox{valign=t}{\frame{\includegraphics[width=.2\textwidth]{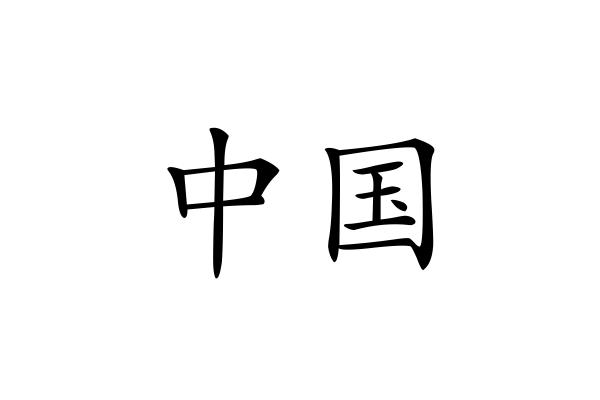}}} & \adjustbox{valign=t}{\frame{\includegraphics[width=.2\textwidth]{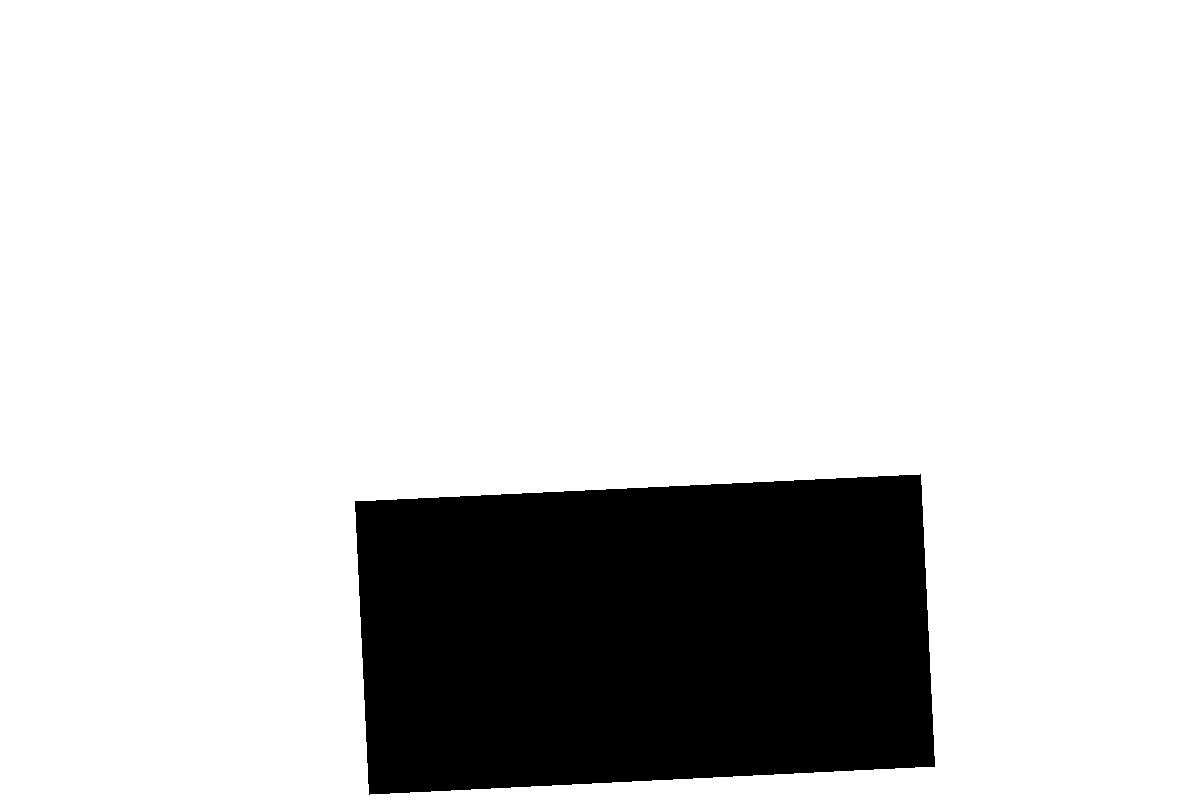}}} & \begin{CJK}{UTF8}{gbsn}一个穿红色衬衫的男人，上面写着中文“中国”\end{CJK} \newline A man in a red shirt with Chinese writing on it "China" \\
        \bottomrule
    \end{tabularx}
    }
    \quad
    \subfloat[An example of our English training data.]{
    \begin{tabularx}{\textwidth}{ccc>{\centering\arraybackslash}X}
        \toprule
        Image & Glyph & Mask & Caption \\
        \midrule
        \adjustbox{valign=t}{\frame{\includegraphics[width=.2\textwidth]{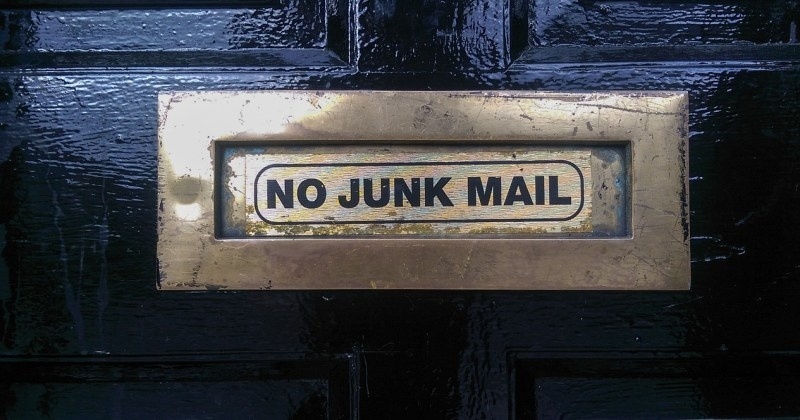}}} & 
        \adjustbox{valign=t}{\frame{\includegraphics[width=.2\textwidth]{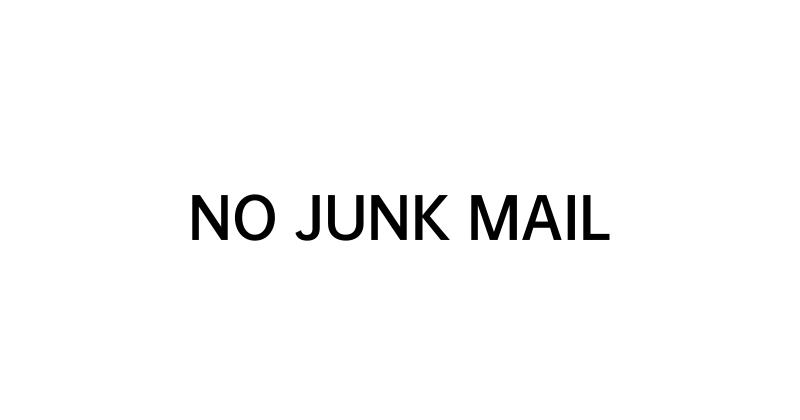}}} & \adjustbox{valign=t}{\frame{\includegraphics[width=.2\textwidth]{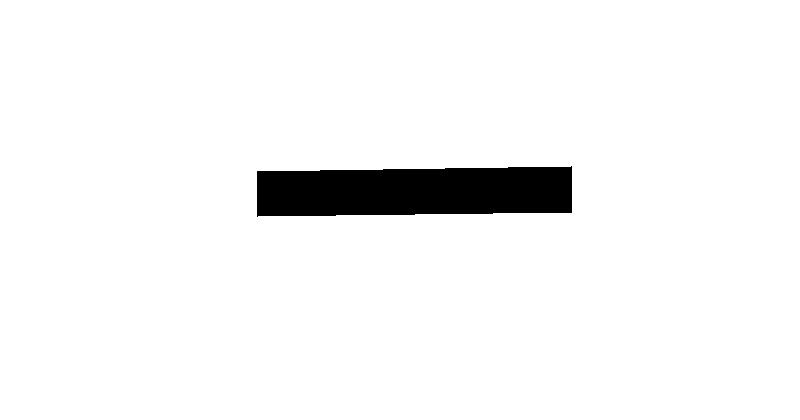}}} & ``no junk mail'' sign on a door \\
        \bottomrule
    \end{tabularx}
    }
    \caption{The constructed text-to-image dataset for Chinese text generation.}
\label{tab:dataset}
\end{table}
We detail the training dataset statistics in Table \ref{tab:dataset}(a) and (b) for Chinese and English, respectively. The Chinese dataset contains 79.2K images with 4800+ unique common Chinese characters and the English dataset consists of 1.93M images with 339k unique English words (including some brand names and websites). Table \ref{tab:dataset}(c) and (d) also show some examples, including the original image, the generated glyph image, the extracted mask and the refined caption.

\section{More experimental results}
\label{app:visualization}

\subsection{More visual text rendering results}
\label{app:more_visualizations}
\begin{figure}
\centering
\includegraphics[width=0.95\textwidth]{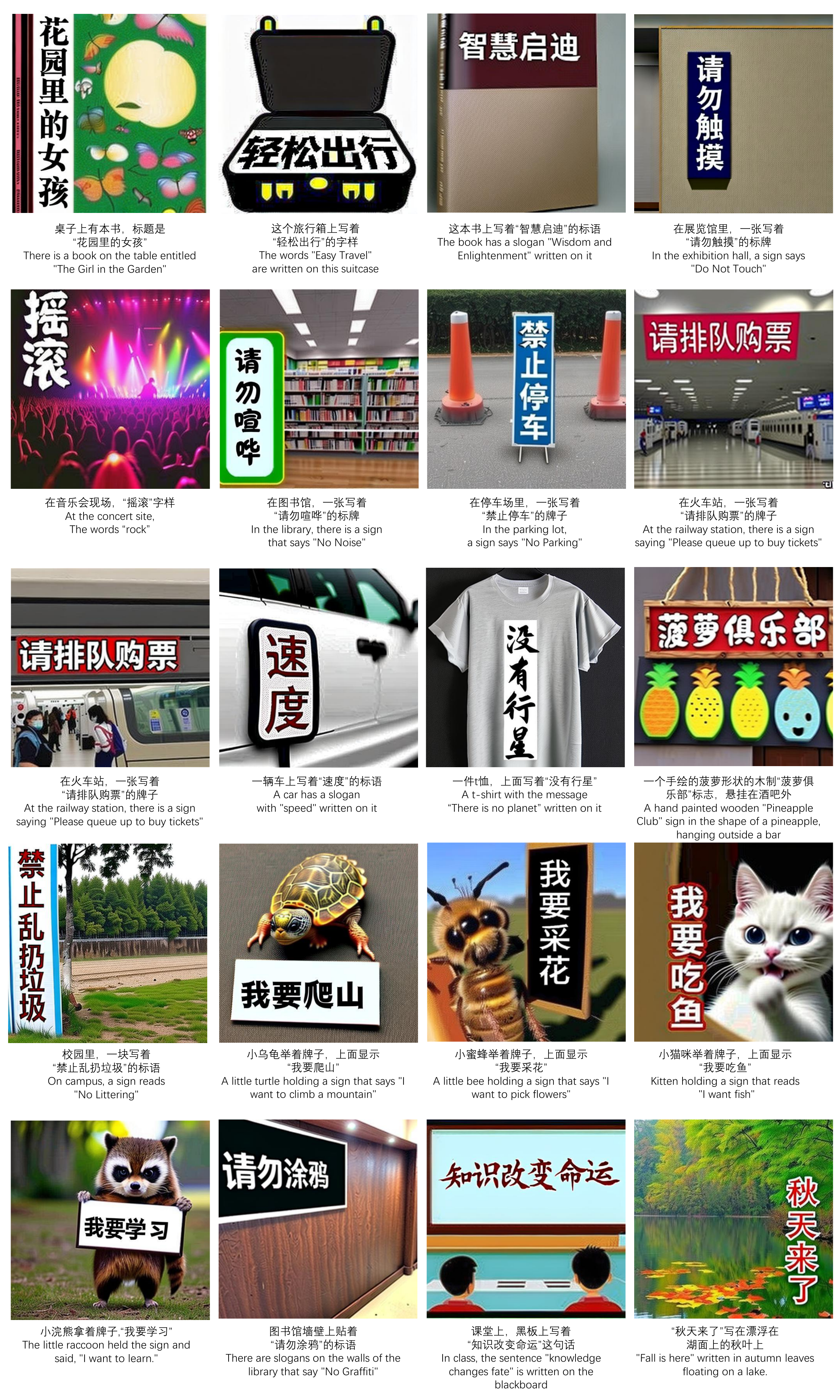}
\caption{More Chinese examples generated using our proposed GlyphDraw.}
\label{fig:more_visual}
\end{figure}

\begin{figure}
\centering
\includegraphics[width=\textwidth]{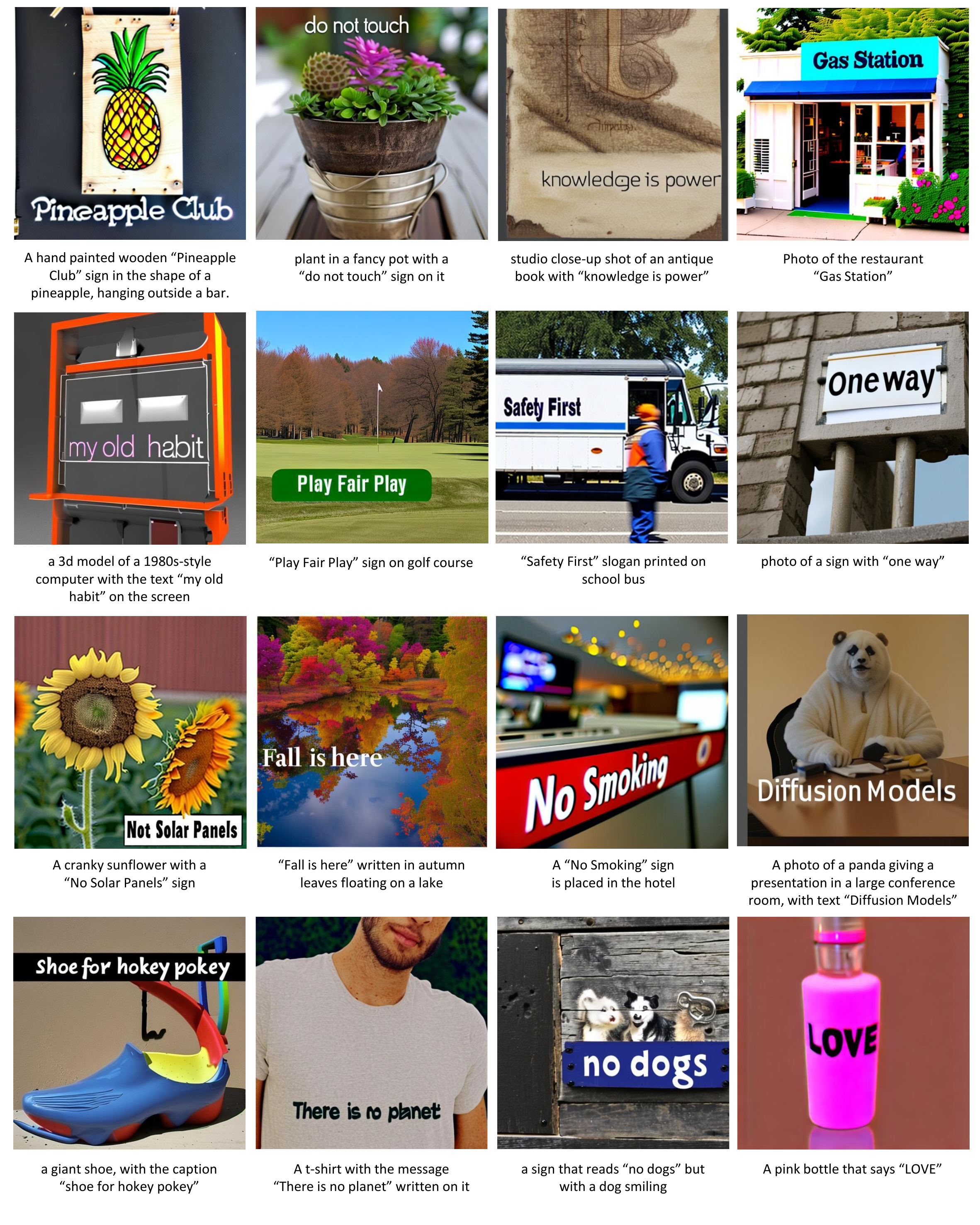}
\caption{More English examples generated using our proposed GlyphDraw.}
\label{fig:more_visual_en}
\end{figure}
More GlyphDraw generated images are provided for demonstration purpose in Fig.~\ref{fig:more_visual} and Fig.~\ref{fig:more_visual_en} for Chinese and English, respectively. We can infer from the visualization results that our Glyphdraw renders accurate characters in generated images as well as blends the characters appropriately into the background of the image using the right font styles and colors. 

\subsection{Synthesized images without visual text}
\label{app:images_no_text}
\begin{table}[htb]
    \centering
    \begin{tabularx}{\textwidth}{>{\centering\arraybackslash\hsize=\hsize}X>    {\centering\arraybackslash\hsize=\hsize}X>{\centering\arraybackslash\hsize=\hsize}X>{\centering\arraybackslash\hsize=\hsize}X>{\centering\arraybackslash\hsize=\hsize}X>{\centering\arraybackslash\hsize=\hsize}X}
        \adjustbox{valign=t}{\frame{\includegraphics[width=.165\textwidth]{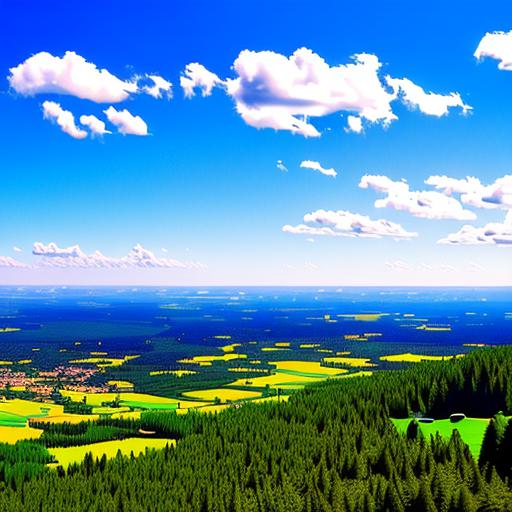}}} & 
        \adjustbox{valign=t}{\frame{\includegraphics[width=.165\textwidth]{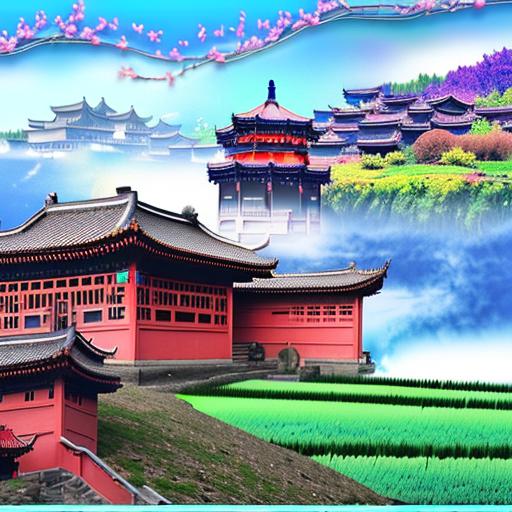}}} & 
        \adjustbox{valign=t}{\frame{\includegraphics[width=.165\textwidth]{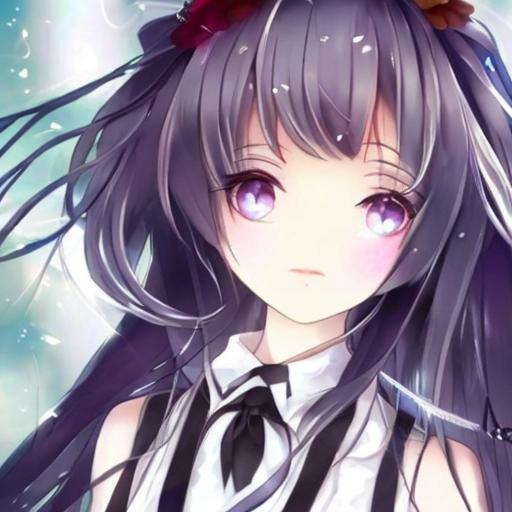}}} & 
        \adjustbox{valign=t}{\frame{\includegraphics[width=.165\textwidth]{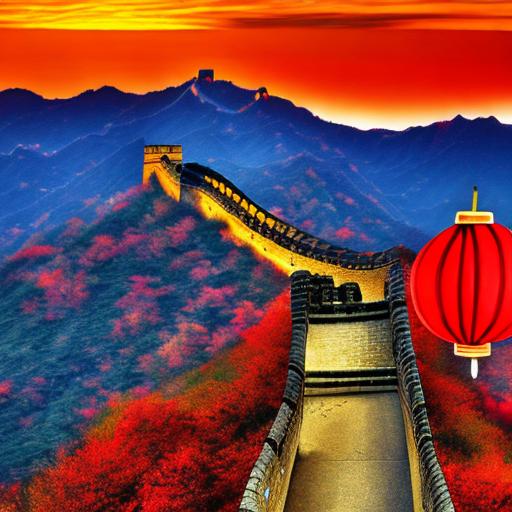}}} & 
        \adjustbox{valign=t}{\frame{\includegraphics[width=.165\textwidth]{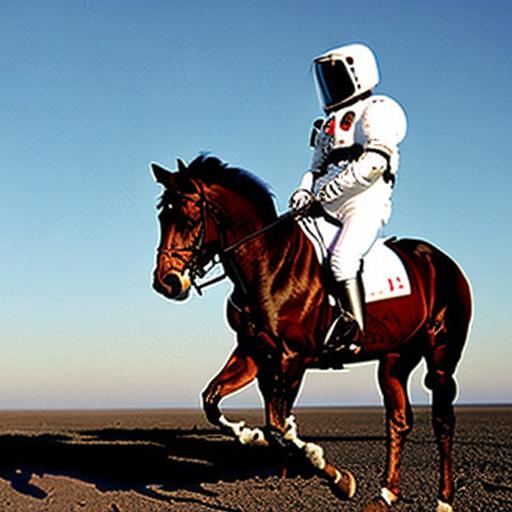}}} & 
        \adjustbox{valign=t}{\frame{\includegraphics[width=.165\textwidth]{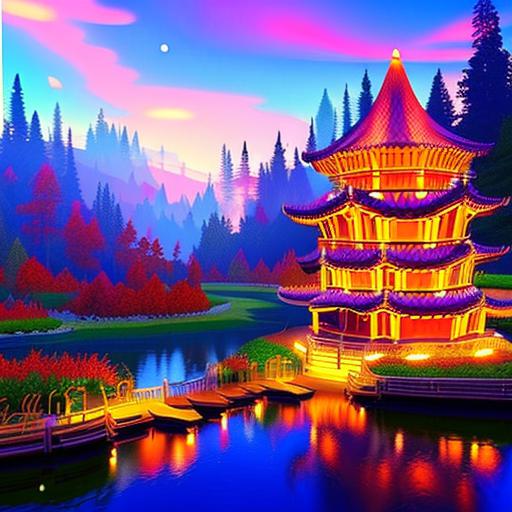}}}  \\
    \end{tabularx}
    \vspace{3mm}
    \caption{Examples of generic images containing no text created by GlyphDraw.}
\label{tab:no_forget}
\end{table}
Table \ref{tab:no_forget} shows some generated images without visual text, revealing that our proposed GlyphDraw can maintain general image synthesis ability. By employing parameter-efficient fine-tuning strategy, our approach achieves slight FID performance degradation as presented in Table \ref{tab:experiment_stats} (1.02 for Chinese model and 1.33 for English model).

\subsection{Visualization for comparisons}
\label{app:visualization_comp}
\begin{table}[htb]
    \centering
    \begin{tabularx}{\textwidth}{>{\centering\arraybackslash\hsize=0.9\hsize}X>{\centering\arraybackslash\hsize=\hsize}X>{\centering\arraybackslash\hsize=\hsize}X>{\centering\arraybackslash\hsize=\hsize}X>{\centering\arraybackslash\hsize=\hsize}X>{\centering\arraybackslash\hsize=\hsize}X>{\centering\arraybackslash\hsize=\hsize}X}
        \toprule
        Fine-tuning & 
        \adjustbox{valign=t}{\frame{\includegraphics[width=.14\textwidth]{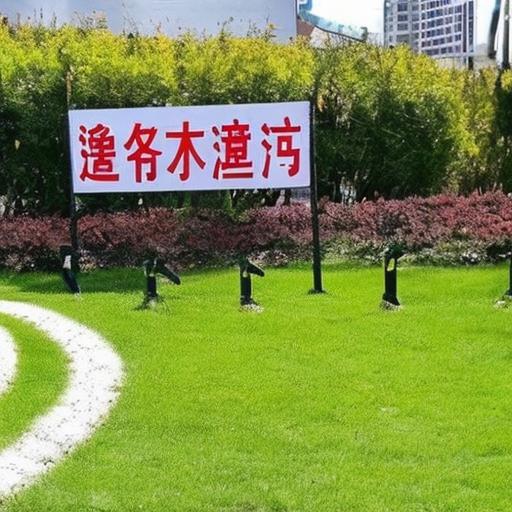}}} & 
        \adjustbox{valign=t}{\frame{\includegraphics[width=.14\textwidth]{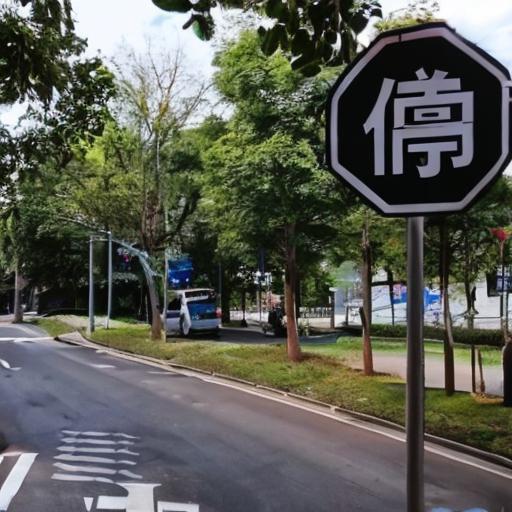}}} & 
        \adjustbox{valign=t}{\frame{\includegraphics[width=.14\textwidth]{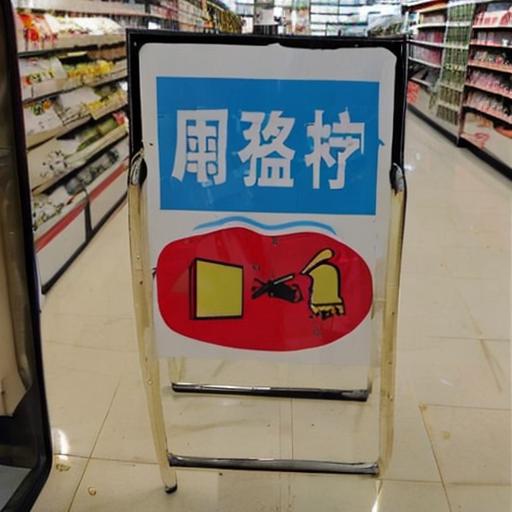}}} & 
        \adjustbox{valign=t}{\frame{\includegraphics[width=.14\textwidth]{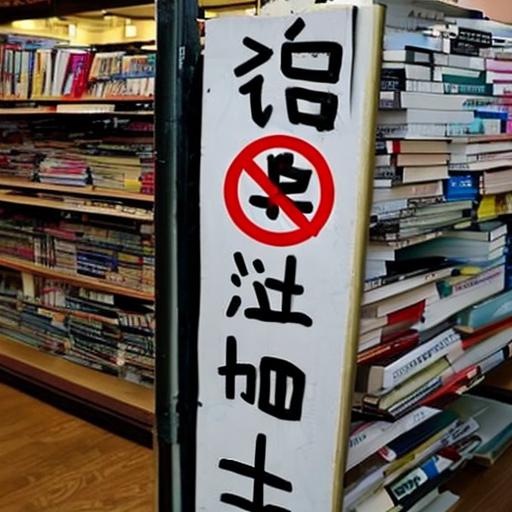}}} & 
        \adjustbox{valign=t}{\frame{\includegraphics[width=.14\textwidth]{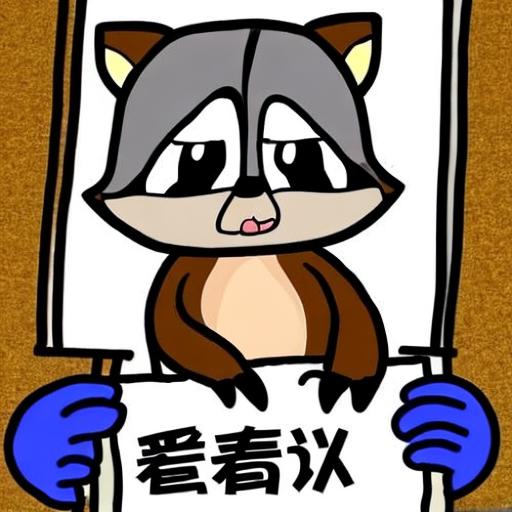}}} & 
        \adjustbox{valign=t}{\frame{\includegraphics[width=.14\textwidth]{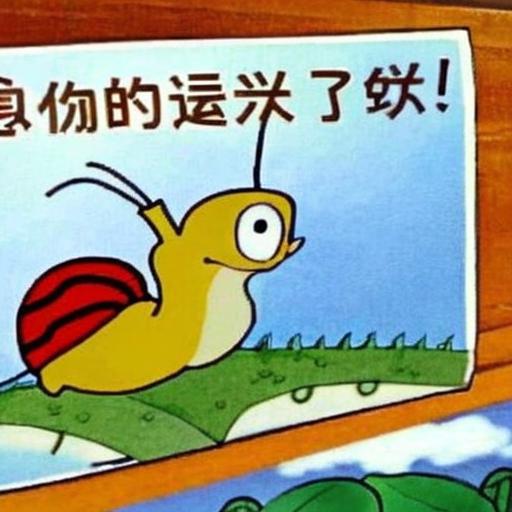}}} \\
      
        ControlNet-Draw & 
        \adjustbox{valign=t}{\frame{\includegraphics[width=.14\textwidth]{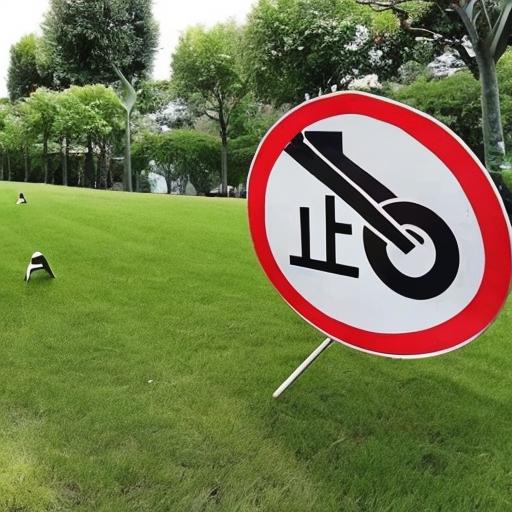}}} & 
        \adjustbox{valign=t}{\frame{\includegraphics[width=.14\textwidth]{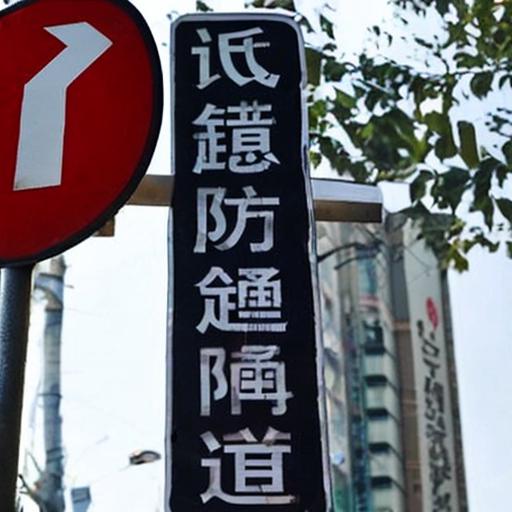}}} & 
        \adjustbox{valign=t}{\frame{\includegraphics[width=.14\textwidth]{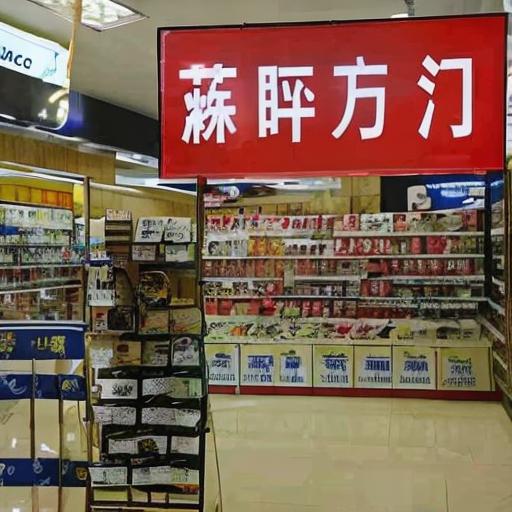}}} & 
        \adjustbox{valign=t}{\frame{\includegraphics[width=.14\textwidth]{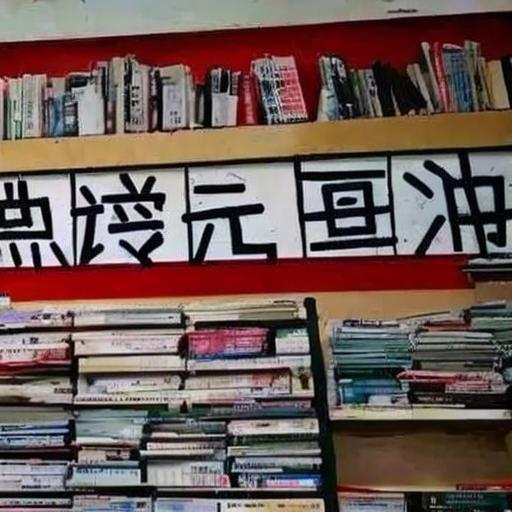}}} & 
        \adjustbox{valign=t}{\frame{\includegraphics[width=.14\textwidth]{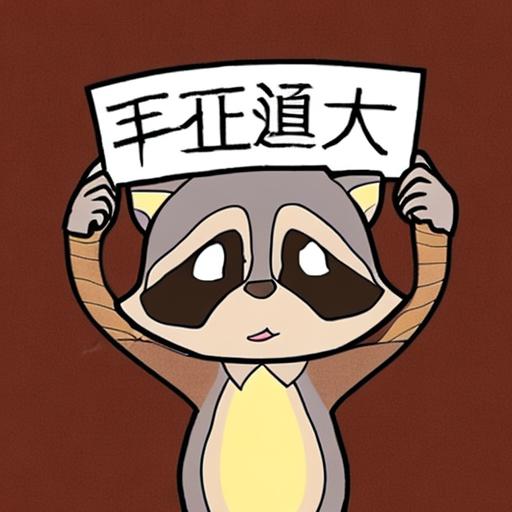}}} & 
        \adjustbox{valign=t}{\frame{\includegraphics[width=.14\textwidth]{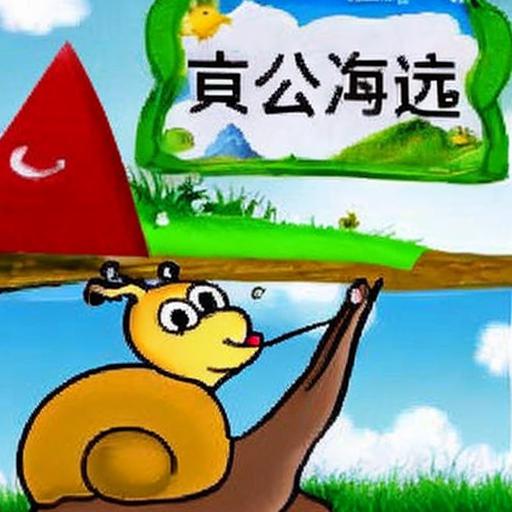}}} \\
      
        GlyphDraw w/o loss weighting & 
        \adjustbox{valign=t}{\frame{\includegraphics[width=.14\textwidth]{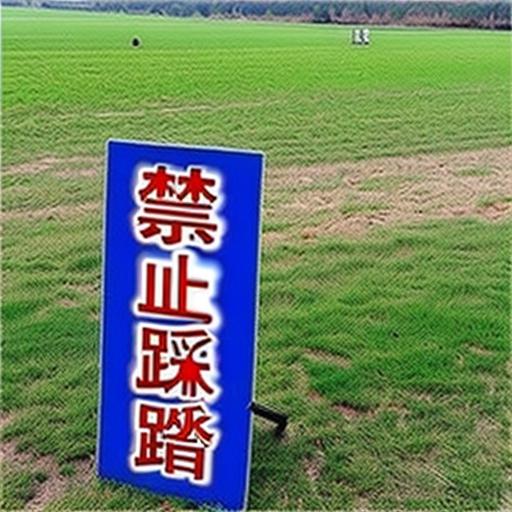}}} & 
        \adjustbox{valign=t}{\frame{\includegraphics[width=.14\textwidth]{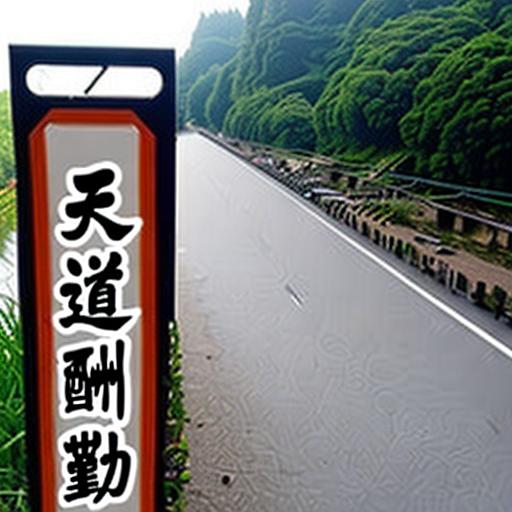}}} & 
        \adjustbox{valign=t}{\frame{\includegraphics[width=.14\textwidth]{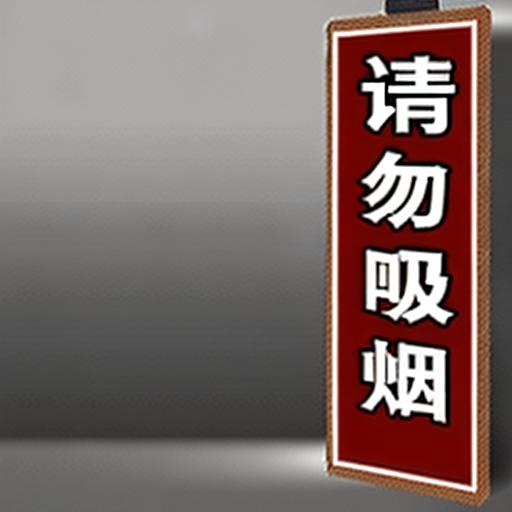}}} & 
        \adjustbox{valign=t}{\frame{\includegraphics[width=.14\textwidth]{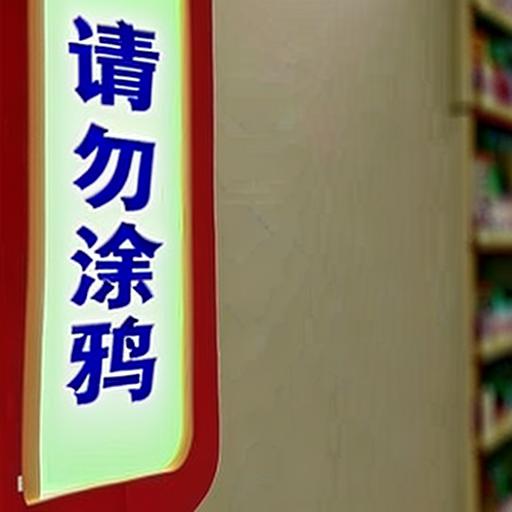}}} & 
        \adjustbox{valign=t}{\frame{\includegraphics[width=.14\textwidth]{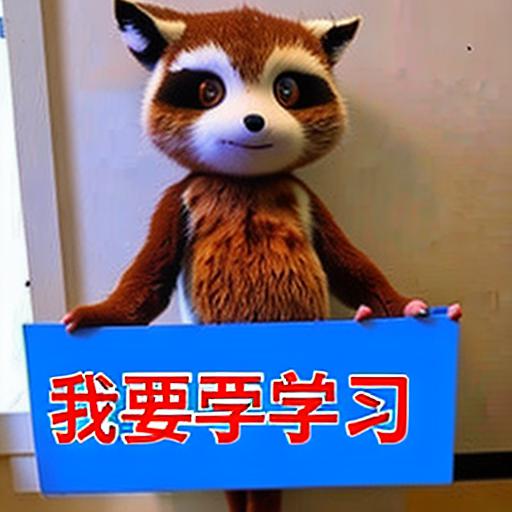}}} & 
        \adjustbox{valign=t}{\frame{\includegraphics[width=.14\textwidth]{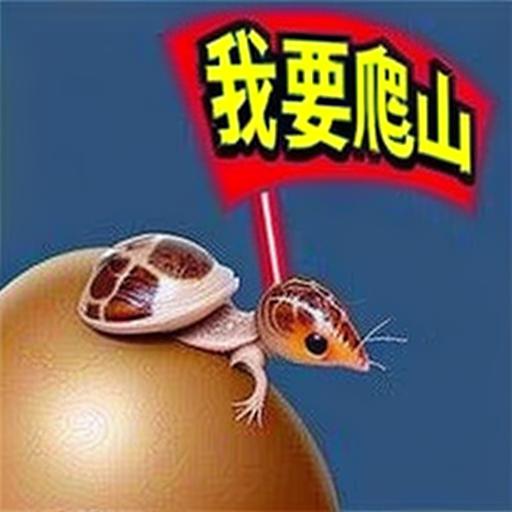}}} \\

        GlyphDraw w/o location mask & 
        \adjustbox{valign=t}{\frame{\includegraphics[width=.14\textwidth]{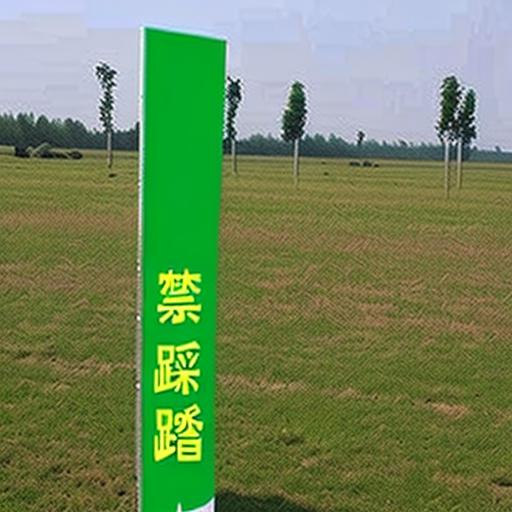}}} & 
        \adjustbox{valign=t}{\frame{\includegraphics[width=.14\textwidth]{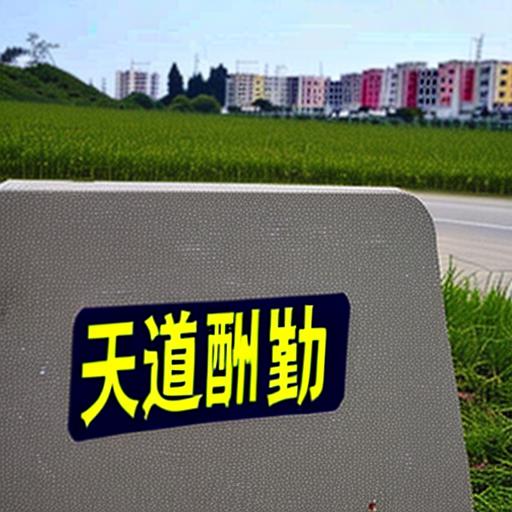}}} & 
        \adjustbox{valign=t}{\frame{\includegraphics[width=.14\textwidth]{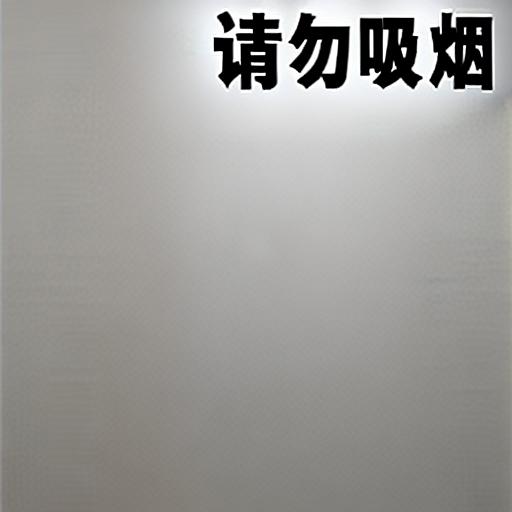}}} & 
        \adjustbox{valign=t}{\frame{\includegraphics[width=.14\textwidth]{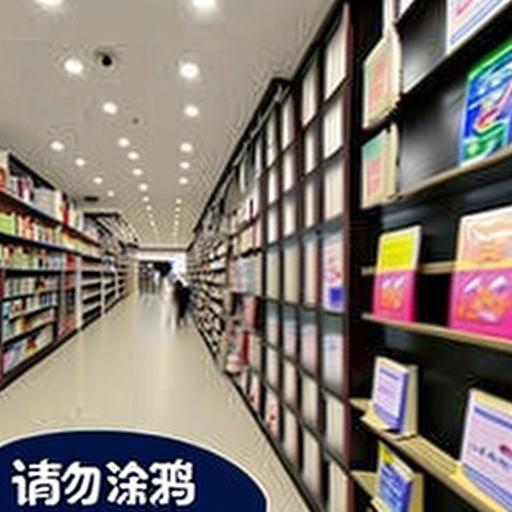}}} & 
        \adjustbox{valign=t}{\frame{\includegraphics[width=.14\textwidth]{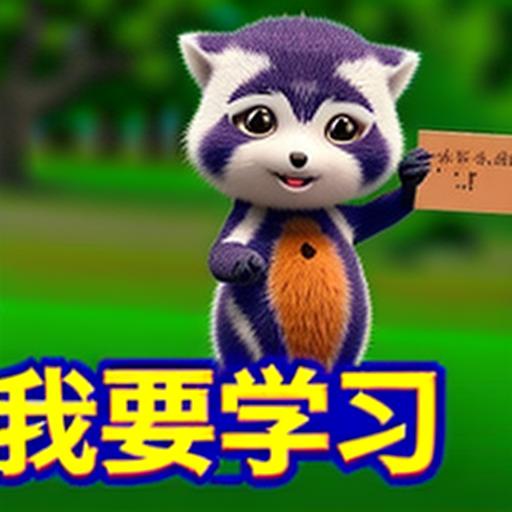}}} & 
        \adjustbox{valign=t}{\frame{\includegraphics[width=.14\textwidth]{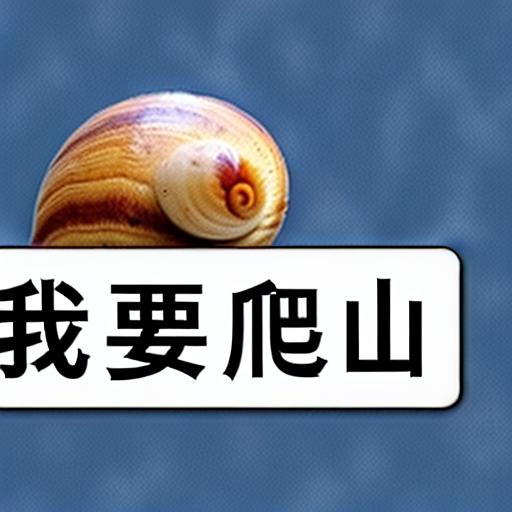}}} \\

        GlyphDraw w/ random mask & 
        \adjustbox{valign=t}{\frame{\includegraphics[width=.14\textwidth]{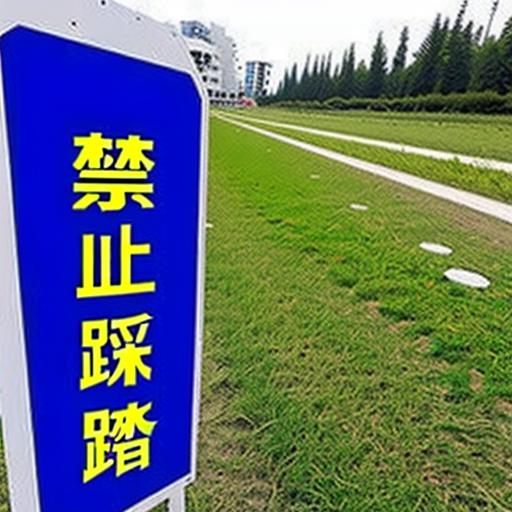}}} & 
        \adjustbox{valign=t}{\frame{\includegraphics[width=.14\textwidth]{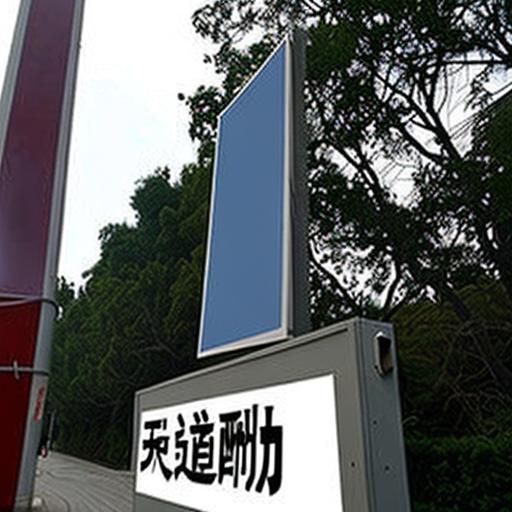}}} & 
        \adjustbox{valign=t}{\frame{\includegraphics[width=.14\textwidth]{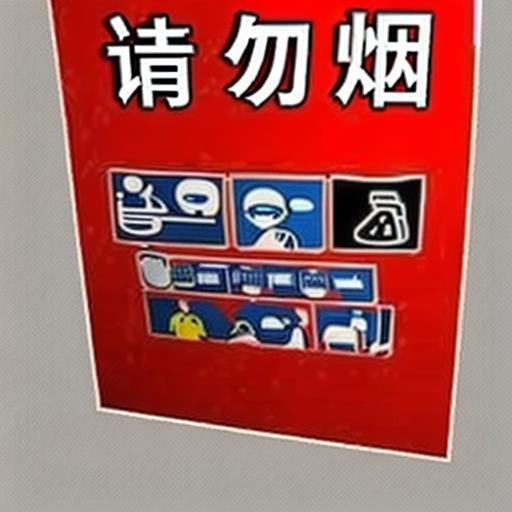}}} & 
        \adjustbox{valign=t}{\frame{\includegraphics[width=.14\textwidth]{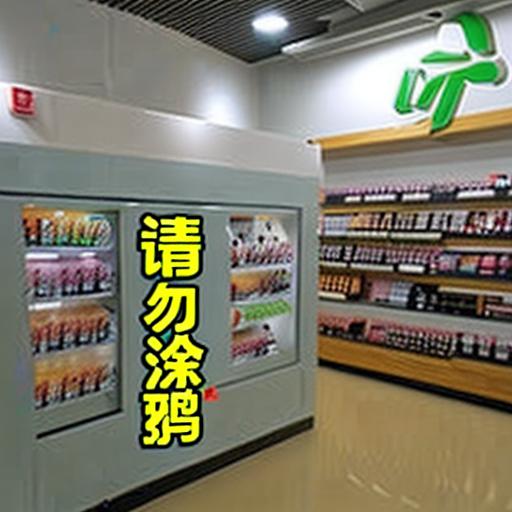}}} & 
        \adjustbox{valign=t}{\frame{\includegraphics[width=.14\textwidth]{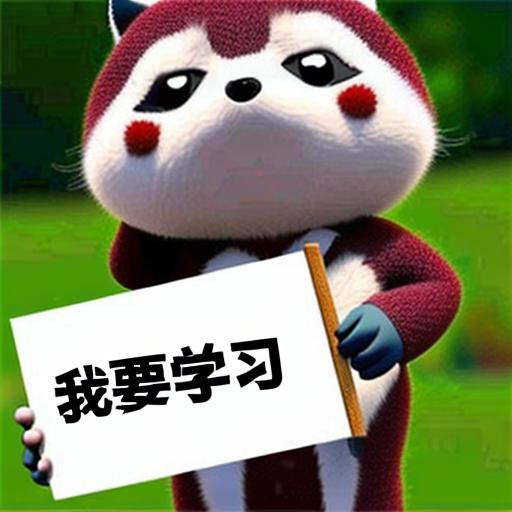}}} & 
        \adjustbox{valign=t}{\frame{\includegraphics[width=.14\textwidth]{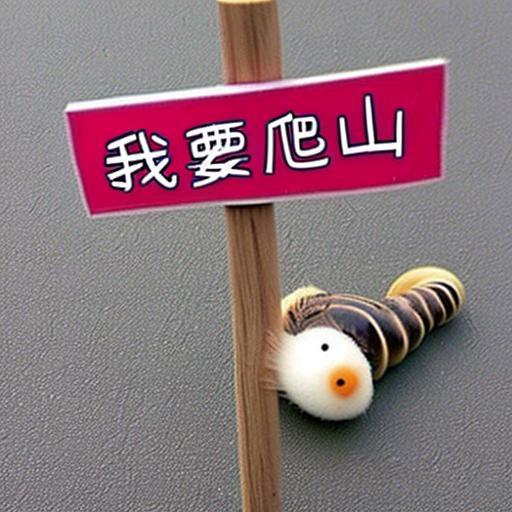}}} \\

        GlyphDraw w/o glyph latent & 
        \adjustbox{valign=t}{\frame{\includegraphics[width=.14\textwidth]{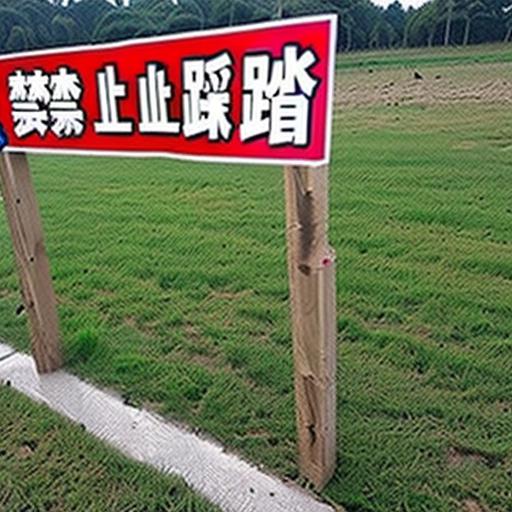}}} & 
        \adjustbox{valign=t}{\frame{\includegraphics[width=.14\textwidth]{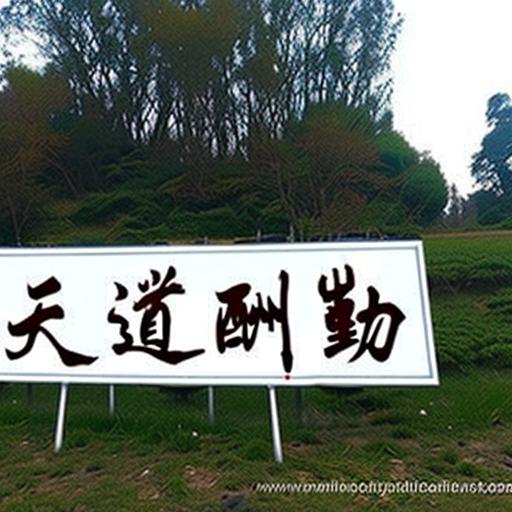}}} & 
        \adjustbox{valign=t}{\frame{\includegraphics[width=.14\textwidth]{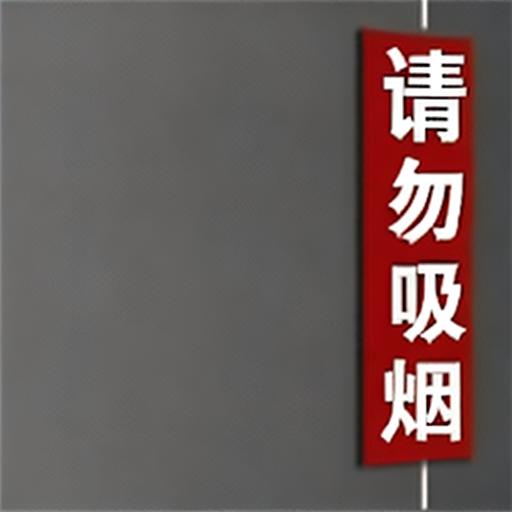}}} & 
        \adjustbox{valign=t}{\frame{\includegraphics[width=.14\textwidth]{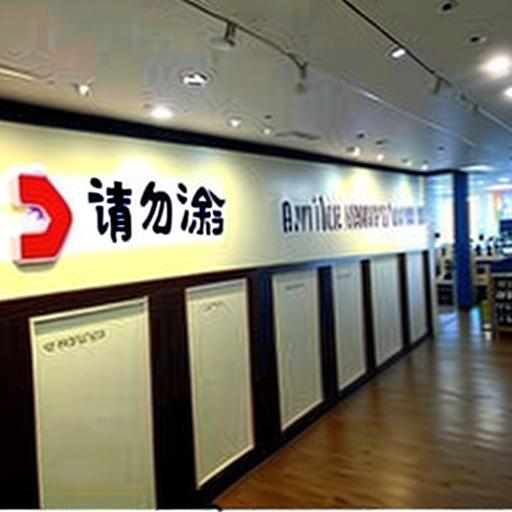}}} & 
        \adjustbox{valign=t}{\frame{\includegraphics[width=.14\textwidth]{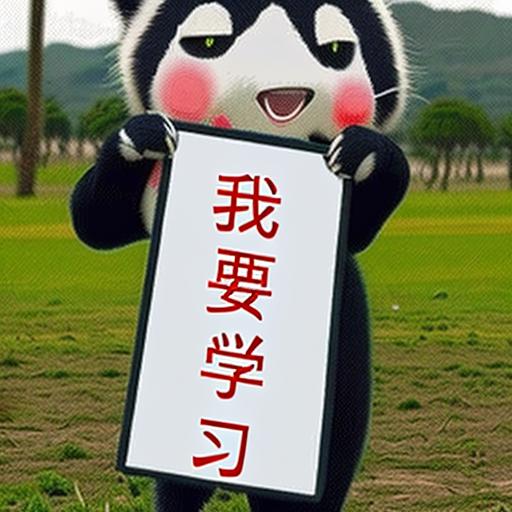}}} & 
        \adjustbox{valign=t}{\frame{\includegraphics[width=.14\textwidth]{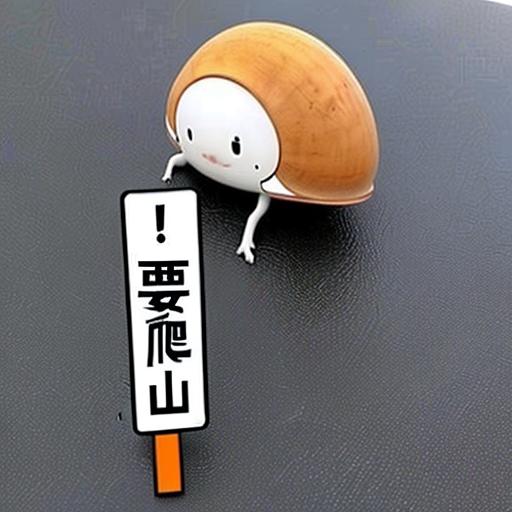}}} \\

        GlyphDraw & 
        \adjustbox{valign=t}{\frame{\includegraphics[width=.14\textwidth]{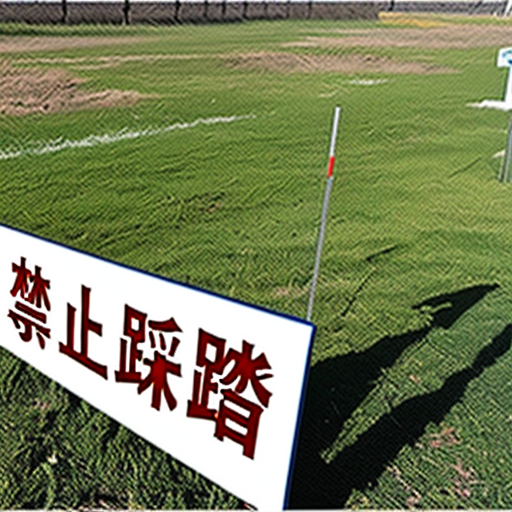}}} & 
        \adjustbox{valign=t}{\frame{\includegraphics[width=.14\textwidth]{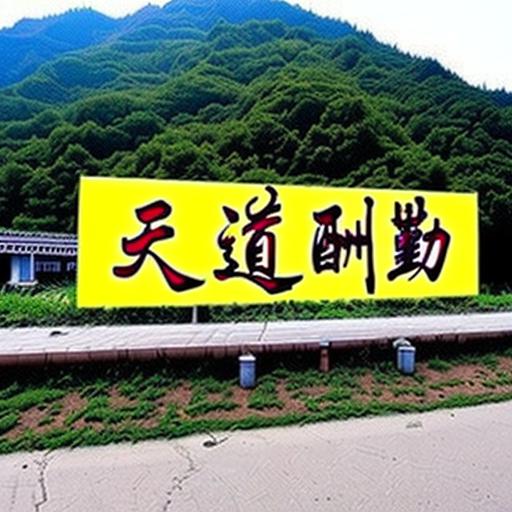}}} & 
        \adjustbox{valign=t}{\frame{\includegraphics[width=.14\textwidth]{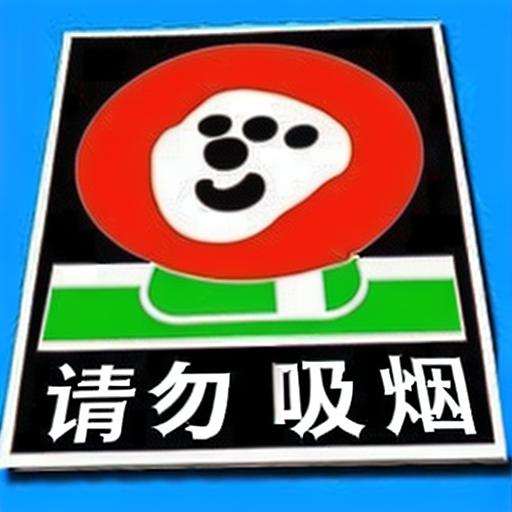}}} & 
        \adjustbox{valign=t}{\frame{\includegraphics[width=.14\textwidth]{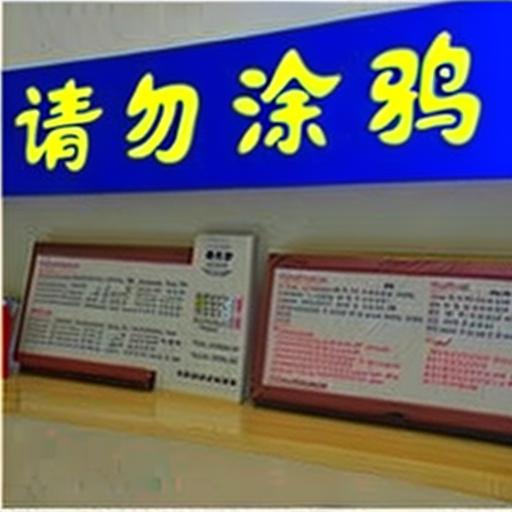}}} & 
        \adjustbox{valign=t}{\frame{\includegraphics[width=.14\textwidth]{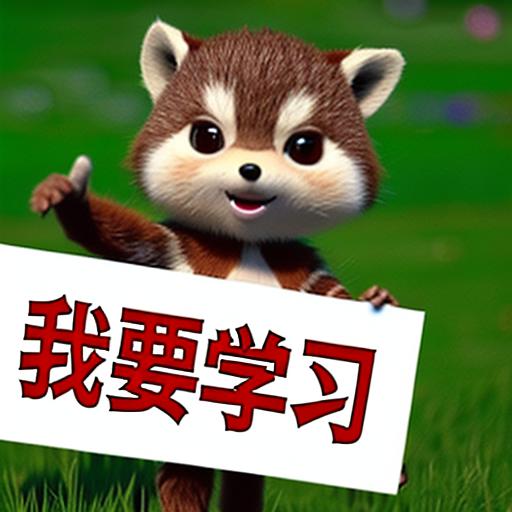}}} & 
        \adjustbox{valign=t}{\frame{\includegraphics[width=.14\textwidth]{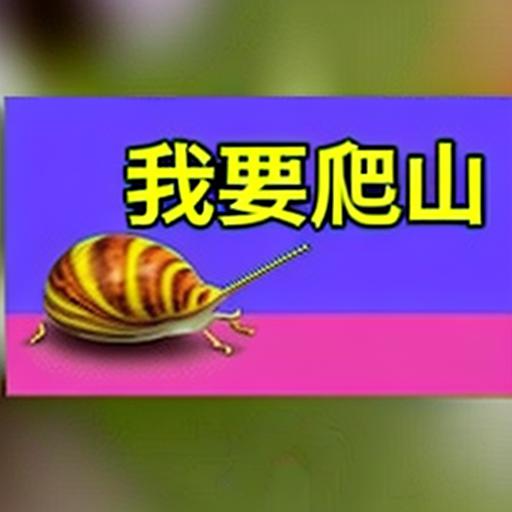}}} \\

        \midrule
        Generated characters & 
        \small\begin{CJK}{UTF8}{gbsn}禁止踩踏\end{CJK} (No trampling) & 
        \small\begin{CJK}{UTF8}{gbsn}天道酬勤\end{CJK} (Heaven rewards diligence)& 
        \small\begin{CJK}{UTF8}{gbsn}请勿吸烟\end{CJK} (No smoking)& 
        \small\begin{CJK}{UTF8}{gbsn}请勿涂鸦\end{CJK} (No graffiti)& 
        \small\begin{CJK}{UTF8}{gbsn}我要学习\end{CJK} (I want to learn)& 
        \small\begin{CJK}{UTF8}{gbsn}我要爬山\end{CJK} (I want to climb mountains)\\
        \bottomrule
    \end{tabularx}
    \vspace{3mm}
    \caption{The visualization results of Chinese text generation, where the images in each row are generated by the same method and the images in each column use the same text prompt as input. }
\label{tab:ablation}
\end{table}

\begin{table}[htb]
    \centering
    \begin{tabularx}{\textwidth}{>{\centering\arraybackslash\hsize=0.9\hsize}X>{\centering\arraybackslash\hsize=\hsize}X>{\centering\arraybackslash\hsize=\hsize}X>{\centering\arraybackslash\hsize=\hsize}X>{\centering\arraybackslash\hsize=\hsize}X>{\centering\arraybackslash\hsize=\hsize}X>{\centering\arraybackslash\hsize=\hsize}X}
        \toprule
        Fine-tuning & 
        \adjustbox{valign=t}{\frame{\includegraphics[width=.14\textwidth]{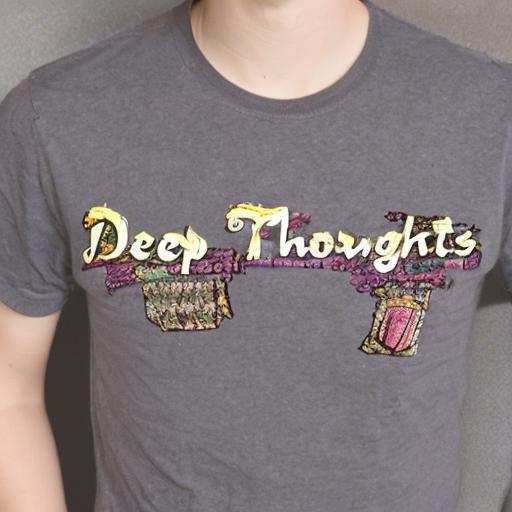}}} & 
        \adjustbox{valign=t}{\frame{\includegraphics[width=.14\textwidth]{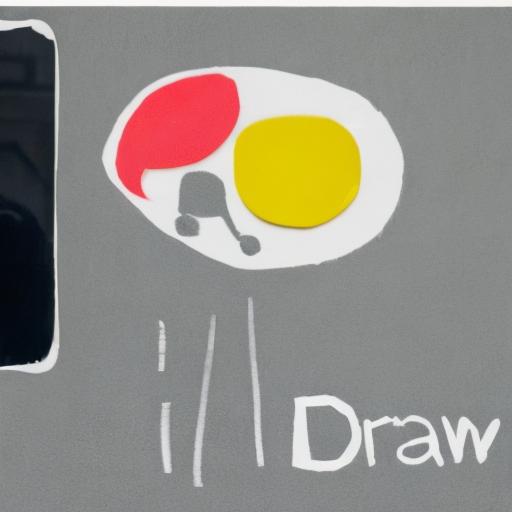}}} & 
        \adjustbox{valign=t}{\frame{\includegraphics[width=.14\textwidth]{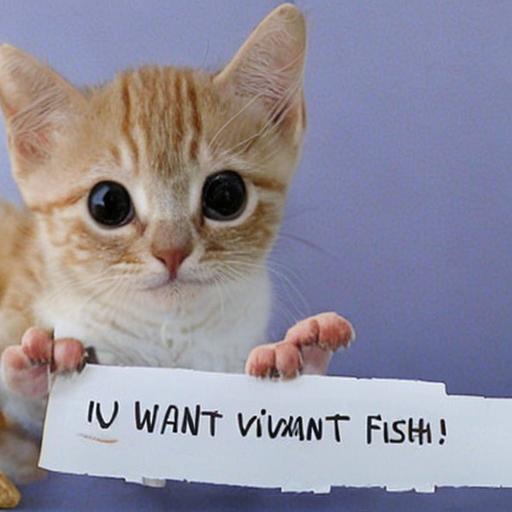}}} & 
        \adjustbox{valign=t}{\frame{\includegraphics[width=.14\textwidth]{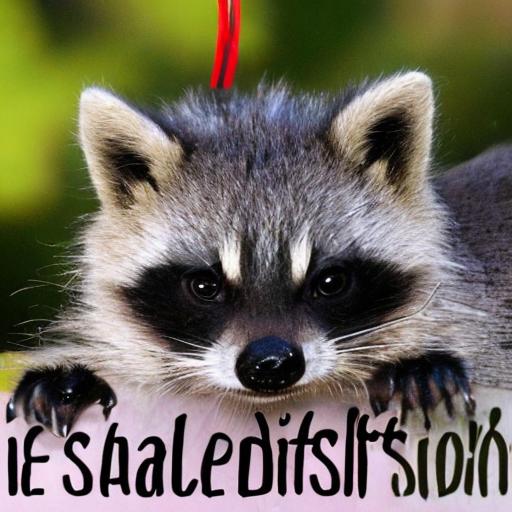}}} & 
        \adjustbox{valign=t}{\frame{\includegraphics[width=.14\textwidth]{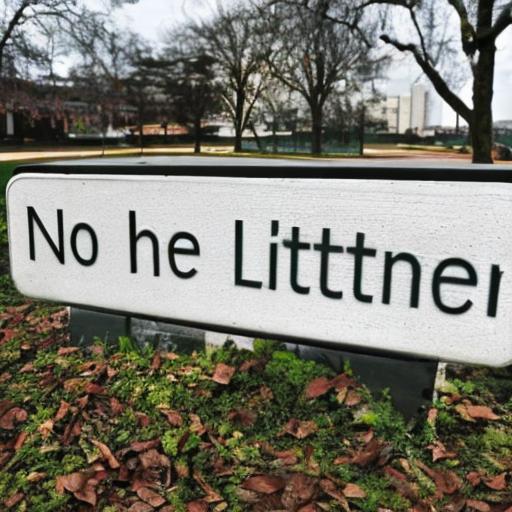}}} & 
        \adjustbox{valign=t}{\frame{\includegraphics[width=.14\textwidth]{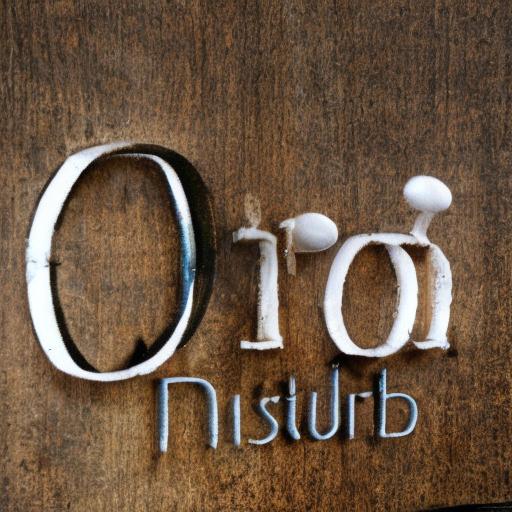}}} \\
      
        ControlNet-Draw & 
        \adjustbox{valign=t}{\frame{\includegraphics[width=.14\textwidth]{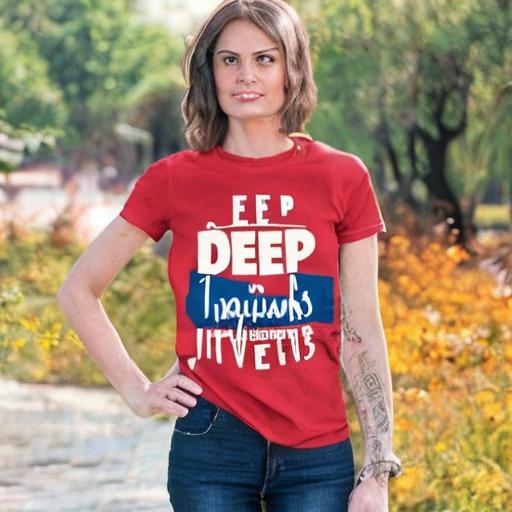}}} & 
        \adjustbox{valign=t}{\frame{\includegraphics[width=.14\textwidth]{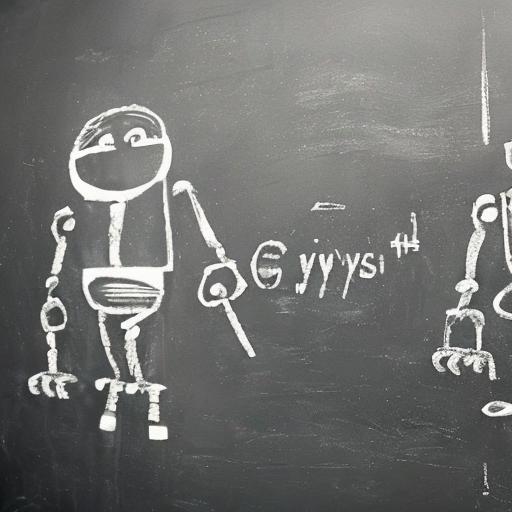}}} & 
        \adjustbox{valign=t}{\frame{\includegraphics[width=.14\textwidth]{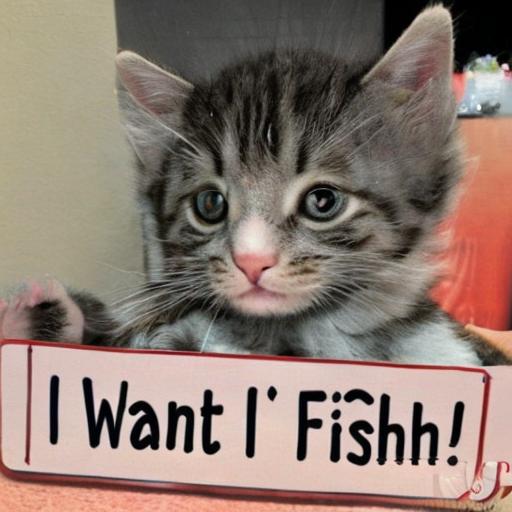}}} & 
        \adjustbox{valign=t}{\frame{\includegraphics[width=.14\textwidth]{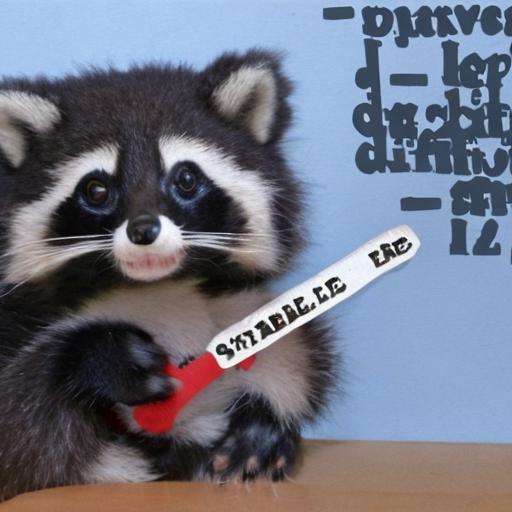}}} & 
        \adjustbox{valign=t}{\frame{\includegraphics[width=.14\textwidth]{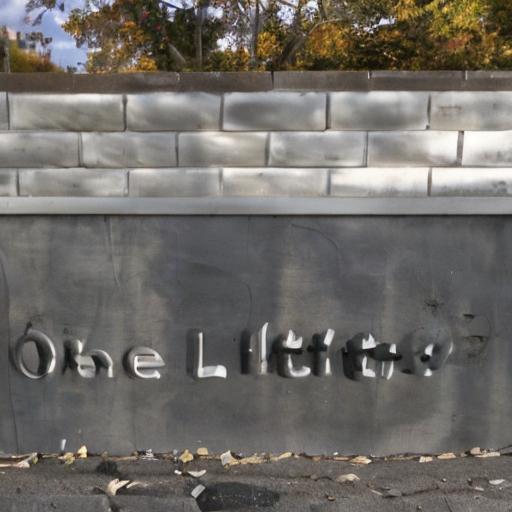}}} & 
        \adjustbox{valign=t}{\frame{\includegraphics[width=.14\textwidth]{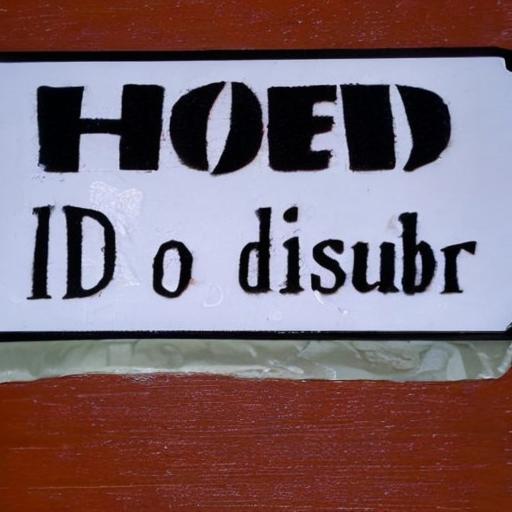}}} \\

        Imagen & 
        \adjustbox{valign=t}{\frame{\includegraphics[width=.14\textwidth]{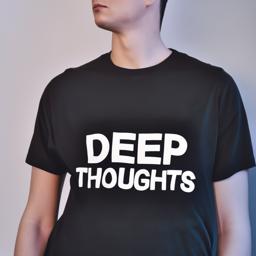}}} & 
        \adjustbox{valign=t}{\frame{\includegraphics[width=.14\textwidth]{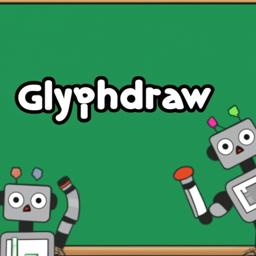}}} & 
        \adjustbox{valign=t}{\frame{\includegraphics[width=.14\textwidth]{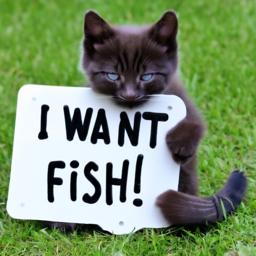}}} & 
        \adjustbox{valign=t}{\frame{\includegraphics[width=.14\textwidth]{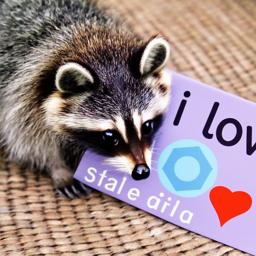}}} & 
        \adjustbox{valign=t}{\frame{\includegraphics[width=.14\textwidth]{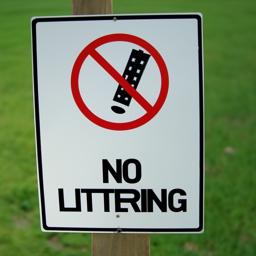}}} & 
        \adjustbox{valign=t}{\frame{\includegraphics[width=.14\textwidth]{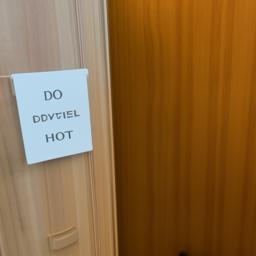}}} \\
      
        GlyphDraw w/o loss weighting & 
        \adjustbox{valign=t}{\frame{\includegraphics[width=.14\textwidth]{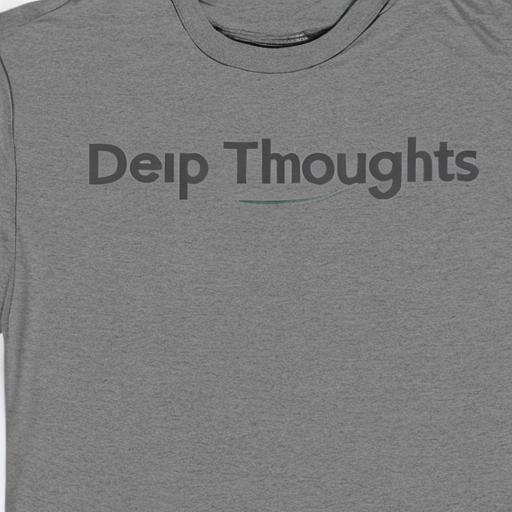}}} & 
        \adjustbox{valign=t}{\frame{\includegraphics[width=.14\textwidth]{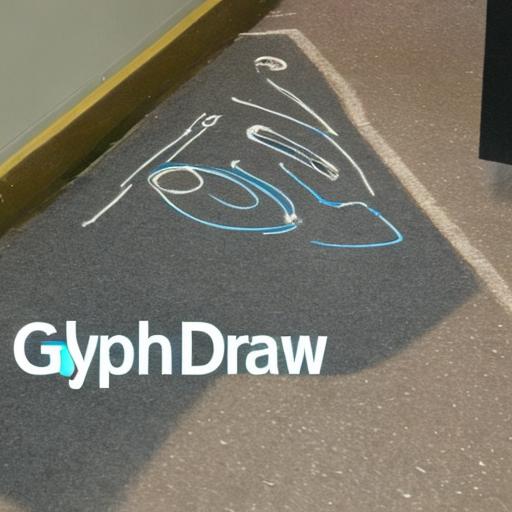}}} & 
        \adjustbox{valign=t}{\frame{\includegraphics[width=.14\textwidth]{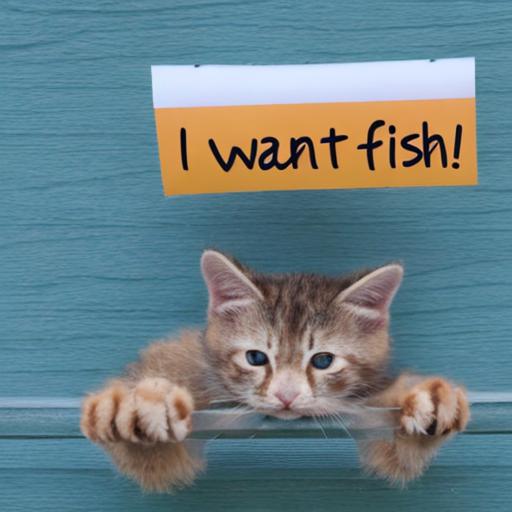}}} & 
        \adjustbox{valign=t}{\frame{\includegraphics[width=.14\textwidth]{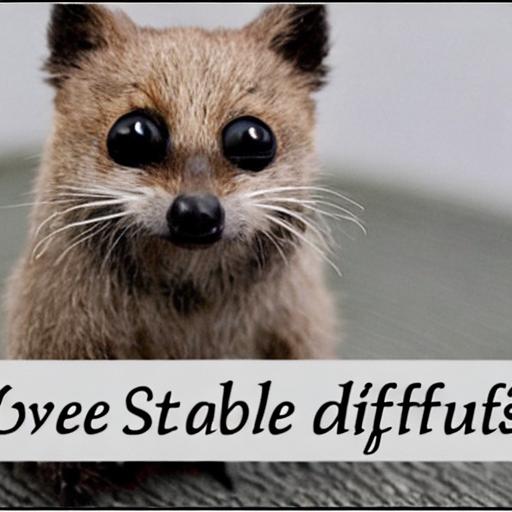}}} & 
        \adjustbox{valign=t}{\frame{\includegraphics[width=.14\textwidth]{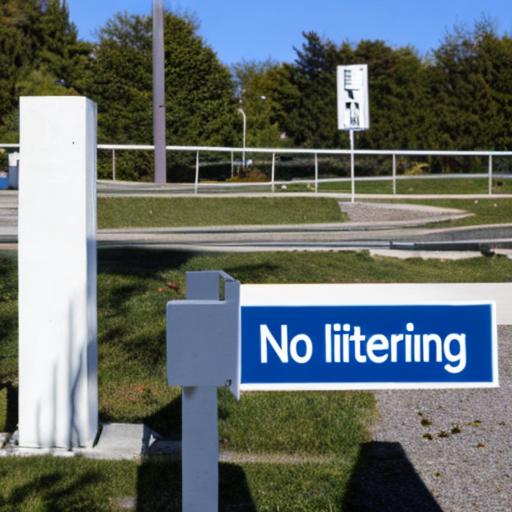}}} & 
        \adjustbox{valign=t}{\frame{\includegraphics[width=.14\textwidth]{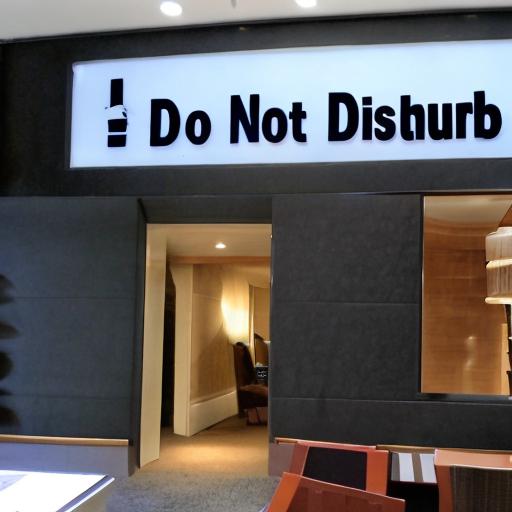}}} \\

        GlyphDraw w/o location mask & 
        \adjustbox{valign=t}{\frame{\includegraphics[width=.14\textwidth]{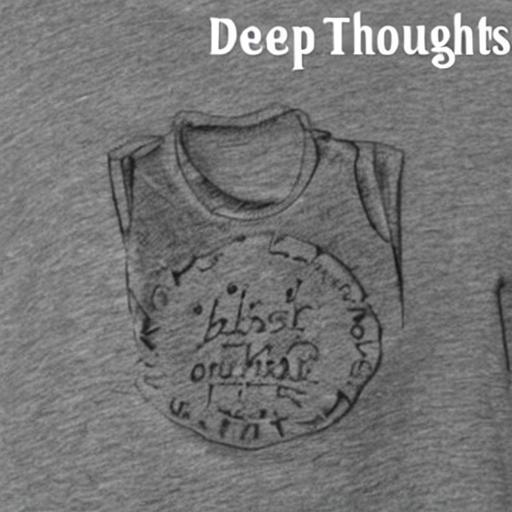}}} & 
        \adjustbox{valign=t}{\frame{\includegraphics[width=.14\textwidth]{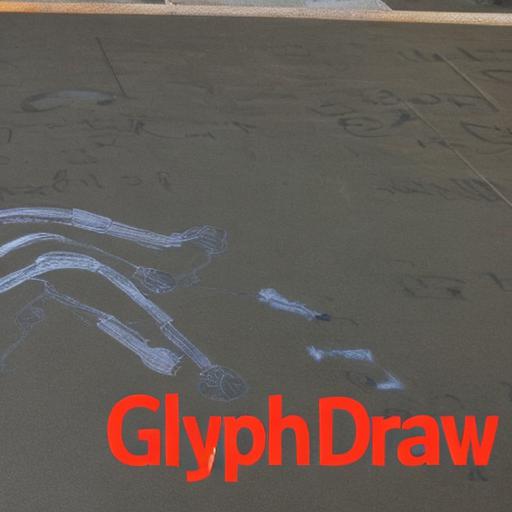}}} & 
        \adjustbox{valign=t}{\frame{\includegraphics[width=.14\textwidth]{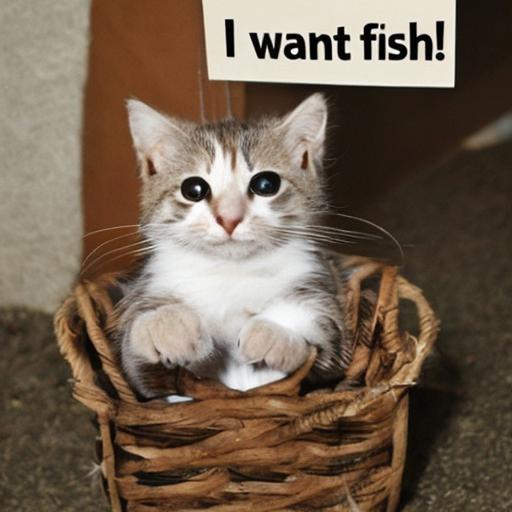}}} & 
        \adjustbox{valign=t}{\frame{\includegraphics[width=.14\textwidth]{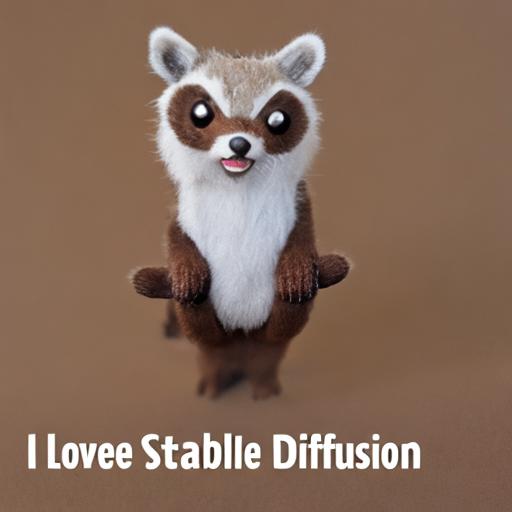}}} & 
        \adjustbox{valign=t}{\frame{\includegraphics[width=.14\textwidth]{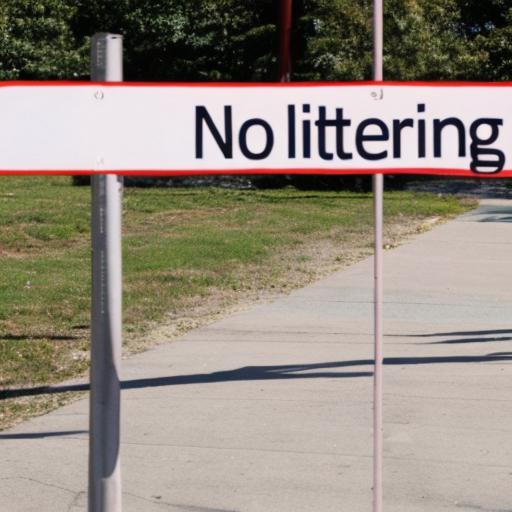}}} & 
        \adjustbox{valign=t}{\frame{\includegraphics[width=.14\textwidth]{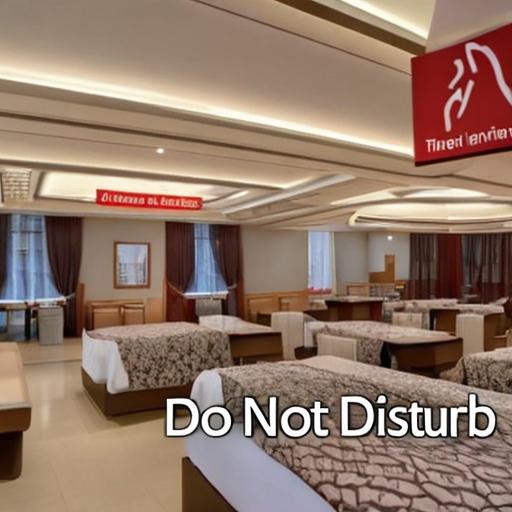}}} \\

        GlyphDraw w/ random mask & 
        \adjustbox{valign=t}{\frame{\includegraphics[width=.14\textwidth]{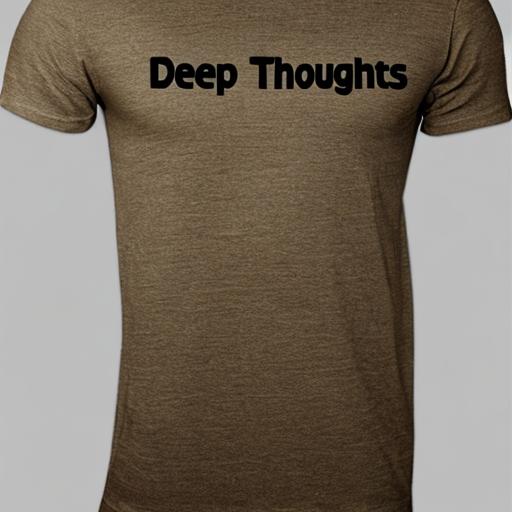}}} & 
        \adjustbox{valign=t}{\frame{\includegraphics[width=.14\textwidth]{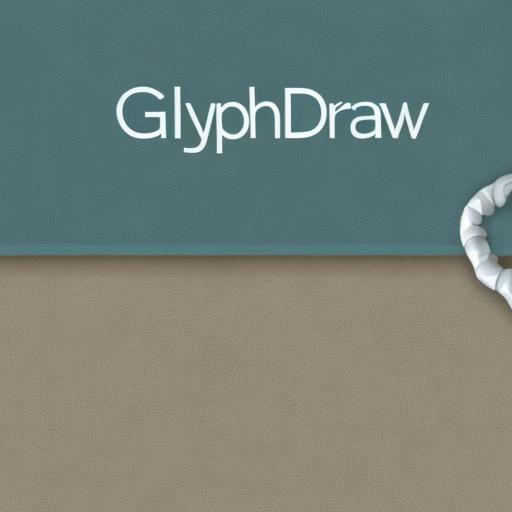}}} & 
        \adjustbox{valign=t}{\frame{\includegraphics[width=.14\textwidth]{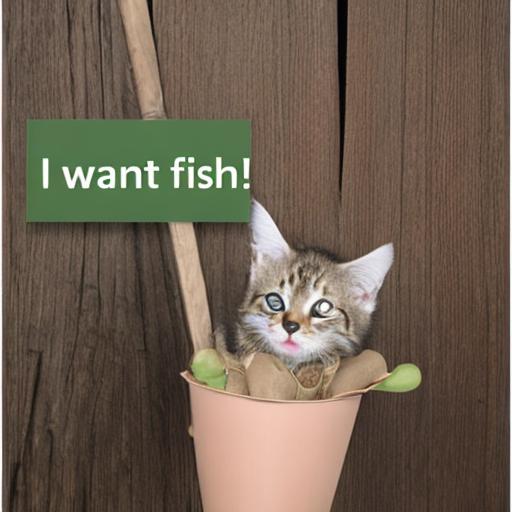}}} & 
        \adjustbox{valign=t}{\frame{\includegraphics[width=.14\textwidth]{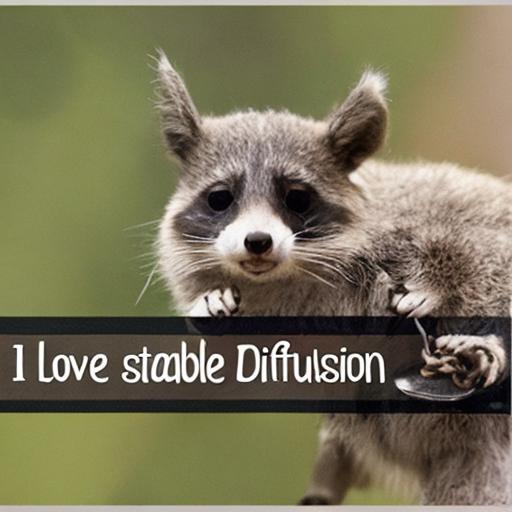}}} & 
        \adjustbox{valign=t}{\frame{\includegraphics[width=.14\textwidth]{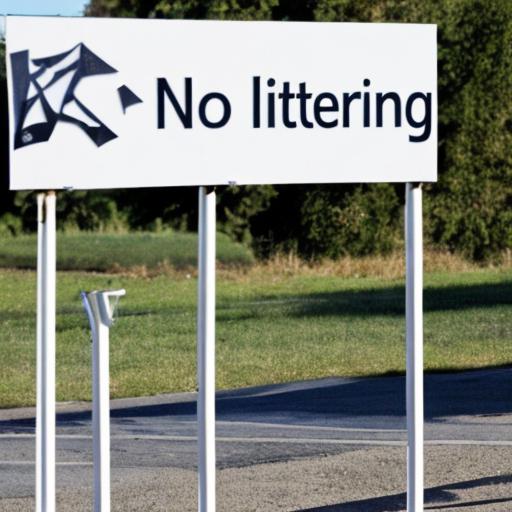}}} & 
        \adjustbox{valign=t}{\frame{\includegraphics[width=.14\textwidth]{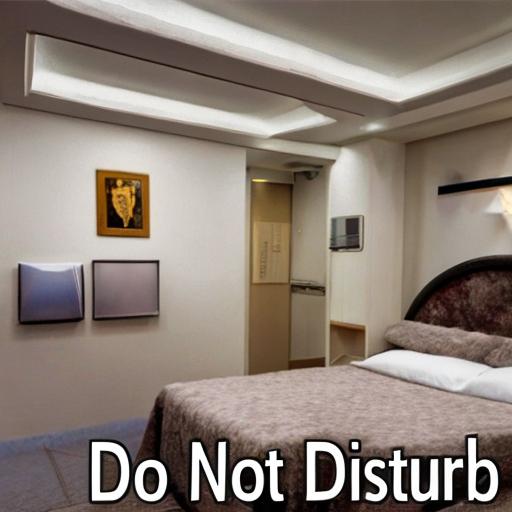}}} \\

        GlyphDraw w/o glyph latent & 
        \adjustbox{valign=t}{\frame{\includegraphics[width=.14\textwidth]{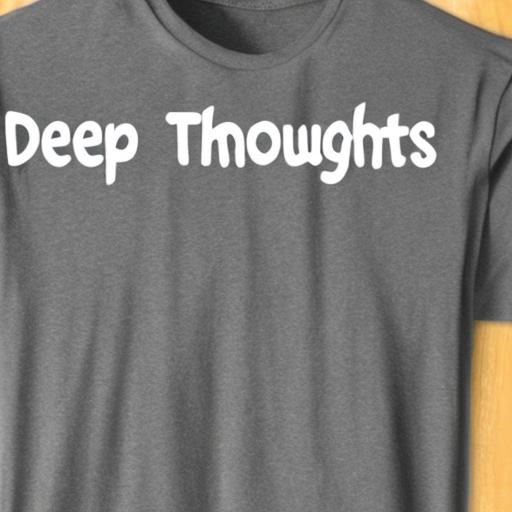}}} & 
        \adjustbox{valign=t}{\frame{\includegraphics[width=.14\textwidth]{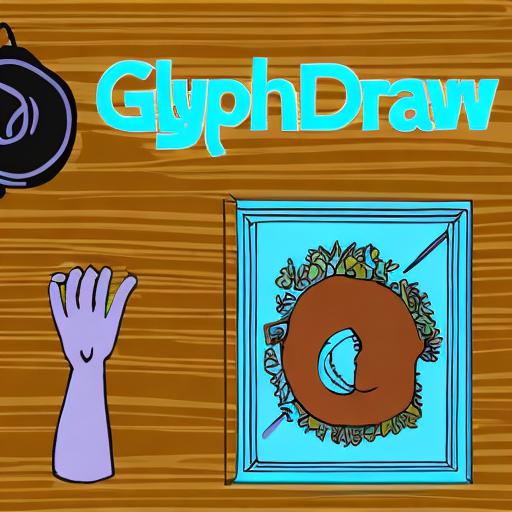}}} & 
        \adjustbox{valign=t}{\frame{\includegraphics[width=.14\textwidth]{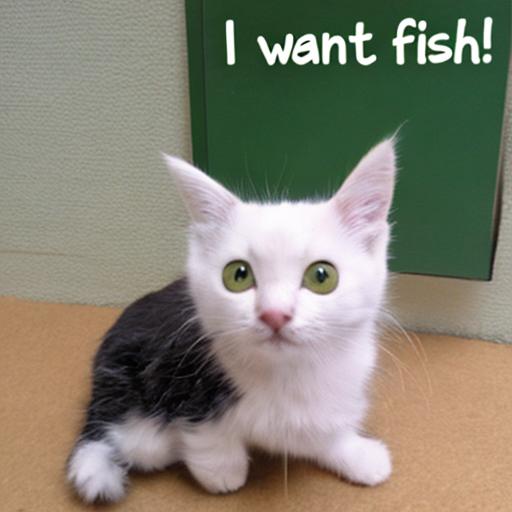}}} & 
        \adjustbox{valign=t}{\frame{\includegraphics[width=.14\textwidth]{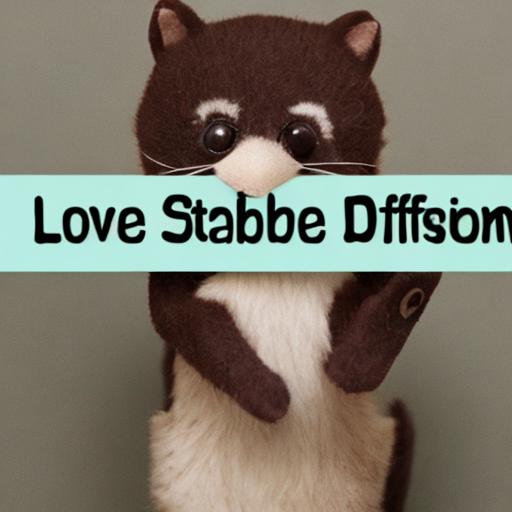}}} & 
        \adjustbox{valign=t}{\frame{\includegraphics[width=.14\textwidth]{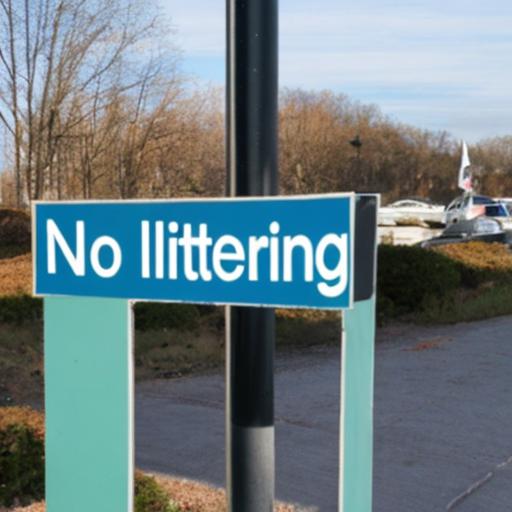}}} & 
        \adjustbox{valign=t}{\frame{\includegraphics[width=.14\textwidth]{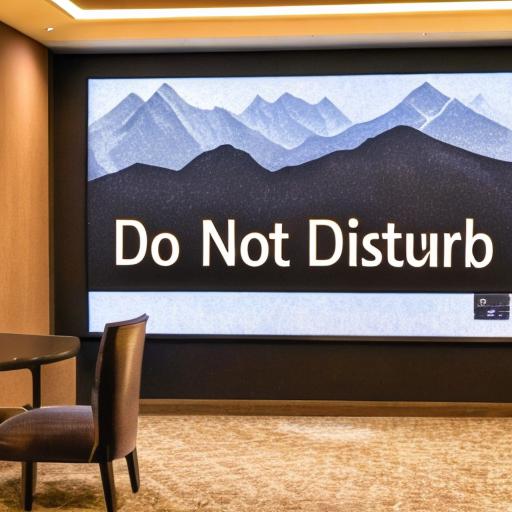}}} \\

        GlyphDraw & 
        \adjustbox{valign=t}{\frame{\includegraphics[width=.14\textwidth]{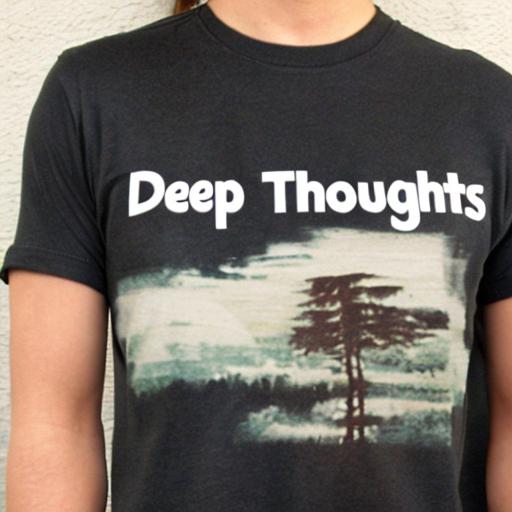}}} & 
        \adjustbox{valign=t}{\frame{\includegraphics[width=.14\textwidth]{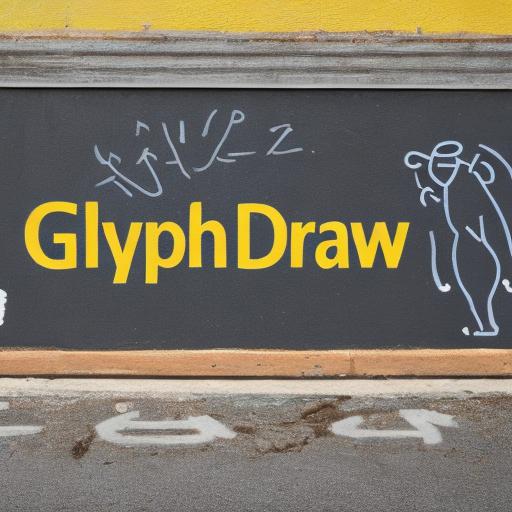}}} & 
        \adjustbox{valign=t}{\frame{\includegraphics[width=.14\textwidth]{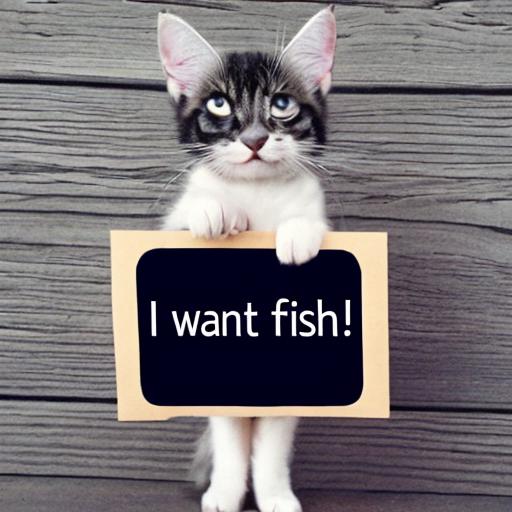}}} & 
        \adjustbox{valign=t}{\frame{\includegraphics[width=.14\textwidth]{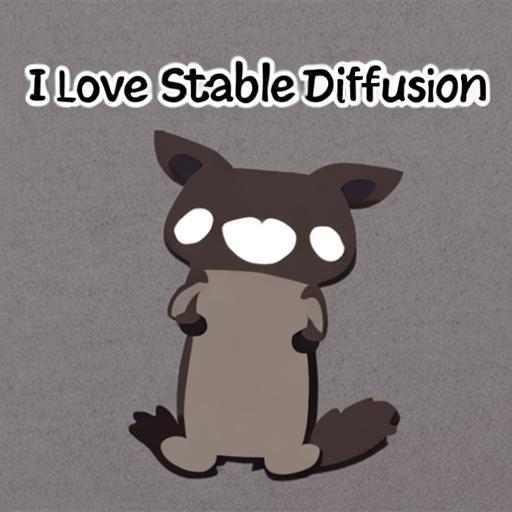}}} & 
        \adjustbox{valign=t}{\frame{\includegraphics[width=.14\textwidth]{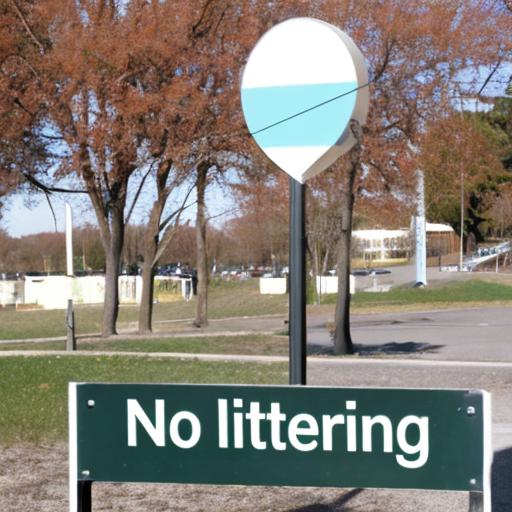}}} & 
        \adjustbox{valign=t}{\frame{\includegraphics[width=.14\textwidth]{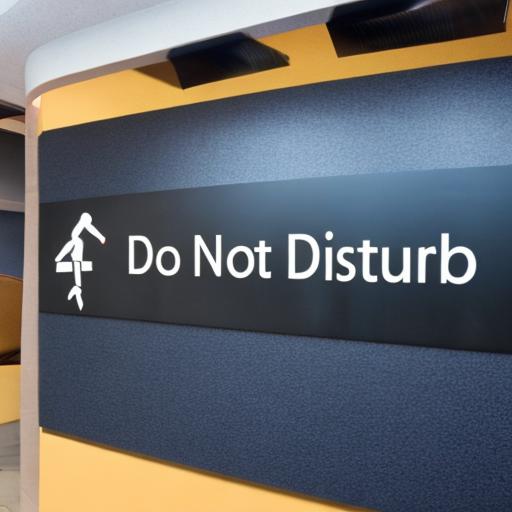}}} \\

        \midrule
        Generated characters & 
        \small Deep Thoughts & 
        \small GlyphDraw & 
        \small I want fish! & 
        \small I Love Stable Diffusion & 
        \small No littering & 
        \small Do Not Disturb \\
        \bottomrule
    \end{tabularx}
    \vspace{3mm}
    \caption{The visualization results of English text generation, where the images in each row are generated by the same method and the images in each column use the same text prompt as input. }
\label{tab:ablation_eng}
\end{table}

Tables \ref{tab:ablation} and \ref{tab:ablation_eng} display some visual text rendering examples generated by employing different methods, including vanilla fine-tuning, ControlNetDraw and different variants of our GlyphDraw for Chinese and English, respectively. Notice that vanilla fine-tuning and ControlNetDraw completely fail to generate correct Chinese characters due to the intricate spatial structures of Chinese glyphs. Even for English, fine-tuning and ControlNetDraw result in wrong or missing letters when rendering visual text. In terms of Imagen, since only the English model is available\footnote{\url{https://github.com/deep-floyd/IF}}, we provide visualization results of English text rendering in Table \ref{tab:ablation_eng} only. As shown, Imagen indeed can generate high-quality images with visual text, but it still struggles to capture all fine-grained character details, leading to incorrect letters or missing words, \textit{e.g.,} the incorrect letter ``y'' in ``GlyphDraw'' and the incomplete word ``love'' in `` I love stable diffusion''. Furthermore, when conducting other ablation studies of different components of our proposed GlyphDraw, we discover that GlyphDraw with the mask prediction module can achieve the generation performance that best conforms to the principle of perspective. Please see the first (``No trampling''), third (``No smoking'') and fourth (``No graffiti'') examples of Table \ref{tab:ablation} and the last example (``Do Not Disturb'') of Table \ref{tab:ablation_eng} for reference. 

\subsection{Accuracy \& FID trade-off}
\label{app:acc_fid_tradeoff}
\begin{figure}[htb]
    \centering
  \includegraphics[width=.5\textwidth]{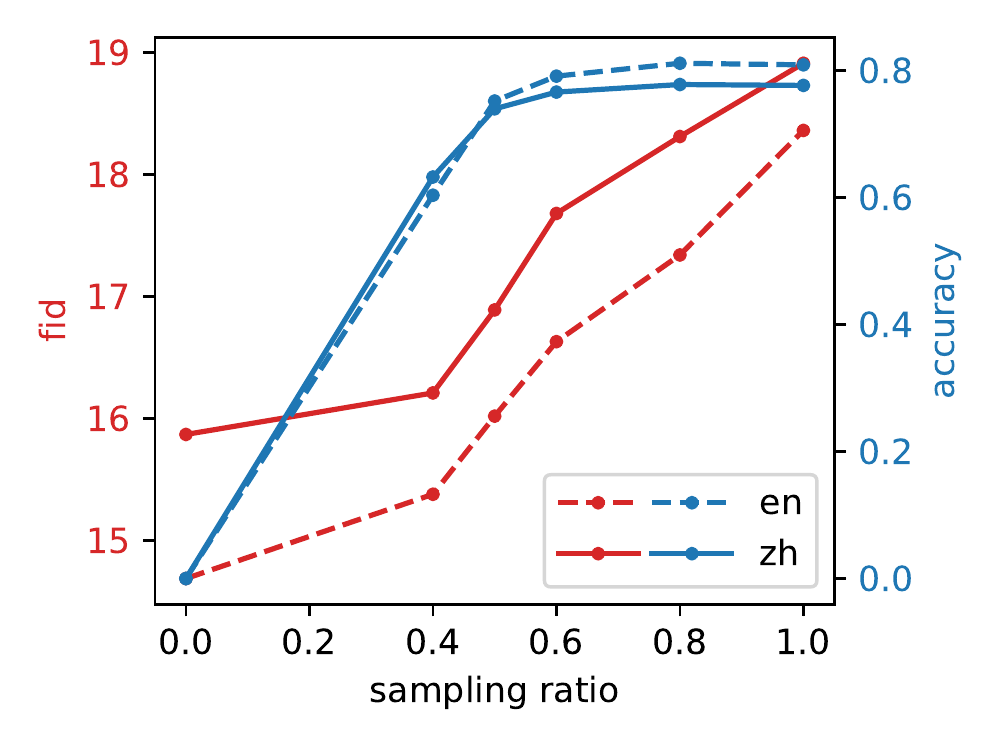}
  \caption{The evaluation results on Chinese and English text generation using different sampling ratios on generation accuracy and FID.}
  \label{fig:fid_acc}
\end{figure}
As proposed in \ref{method:inference}, we introduce a hyper-parameter $r$ to devide the generation steps into two phases, sampling with our GlyphDraw model and the original Stable Diffusion model. We illustrate the diagram of OCR accuracy versus FID performance over different $r$ values for both English and Chinese testsets, as shown in Fig. \ref{fig:fid_acc}. It can be observed that both the FID value and OCR accuracy rise as the sampling ratio $r$ increases. Notice that the accuracies are almost saturated and the FID values remain relatively low when sampling ratio reaches 0.5. In order to make a trade-off between open-domain generalization and visual text rendering ability of the GlyphDraw model, we finally chose $r=0.5$ for all experiments.

\section{Test benchmark details}
\label{app:test_benchmark}

\subsection{OCR evaluation criteria}
We evaluate the text generation accuracy of images produced by image synthesis models using a well-performing OCR tool \cite{du2021pp}. We consider the generation correct if the requested text is included in the OCR recognition results from the generated image and incorrect otherwise. For English text generation evaluation, we ignore letter case.

\subsection{Detailed OCR evaluation results}
\label{app:acc_detail}
\begin{figure}[htb]
\centering
  \includegraphics[width=\textwidth]{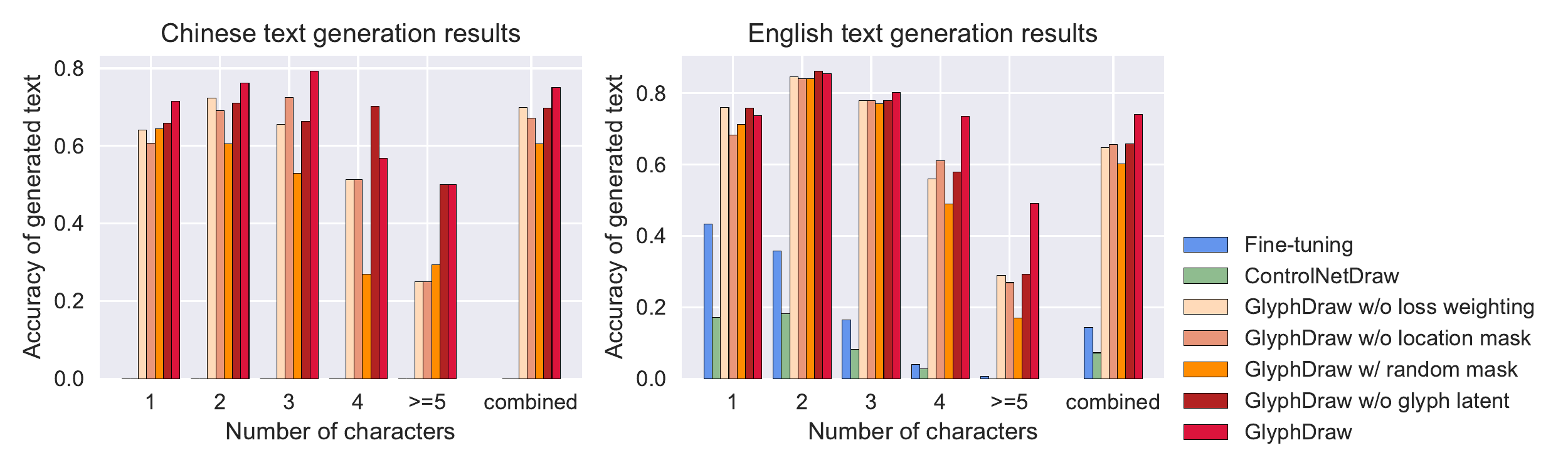}
  \caption{Demonstration of the quality of generated Chinese and English text in terms of accuracy with respect to different number of generated characters, evaluated by an OCR recognition model.}
  \label{fig:acc_detail}
\end{figure}
Fig. \ref{fig:acc_detail} provides specific OCR accuracy for various character lengths for both Chinese and English. Similar to popular belief, OCR accuracy tends to decline as character length rise. Notice that fine-tuning and ControlNetDraw perform extremely poor for both Chinese and English visual text rendering.

\subsection{Test prompts}
\label{app:test_prompts}
Here we detailedly list all the prompts uesd in our ChineseDrawText test benchmark proposed in Sec.~\ref{implem_details_eval} as follows.

\begin{CJK}{UTF8}{gbsn}
1: 街边的路牌，内容显示“天道酬勤”(A street sign on the street reads "Heaven rewards diligence")

2: 海边有一个石碑，上面刻着“北戴河”(Thereis a stone tablet on the
seashore with the inscription "Beidaihe" )

3: 一个粉红色的瓶子，上面写着“爱情”(A pink bottle that says "Love")

4: 桌子上有本书，标题是“花园里的女孩”(There is a book on the table entitled
"The Girl in the Garden")

5: 一张精致的浪漫名片，上面写着“我爱你”(A sophisticated romantic business card that says "I love you")

6: 一只猫在读一本书，书的标题是“捉鼠大法”(A cat is reading a book with the title "How to catch mice")

7: 小浣熊拿着杯子喝水，杯子上印着“浣熊”(A little raccoon drinks from a cup with the word "raccoon" on it)

8: 小浣熊举着牌子,上面显示“我要学习”(The little raccoon held the sign and said, "I want to learn")

9: 小猫咪举着牌子，上面显示“我要吃鱼”(Kitten holding a sign that reads "I want fish")

10: 小熊猫举着牌子，上面写着“我要爬树”(A lesser panda holding a sign that says "I want to climb a tree")

11: 小青蛙举着牌子，上面写着“我要跳舞”(A little frog holding a sign that says "I want to dance")

12: 小海豚举着牌子，上面显示“我要游泳”(A baby dolphin holds a sign that says "I want to swim")

13: 小松鼠举着牌子，上面显示“我要储存粮食”(A little squirrel holds a sign that says "I want to store food")

14: 小乌龟举着牌子，上面显示“我要爬山”(A little turtle holding a sign that says "I want to climb a mountain")

15: 小蚂蚁举着牌子，上面写着“我要搬家”(A little ants holding a sign that says "I want to move house")

16: 小蜜蜂举着牌子，上面显示“我要采花”(A little bee holding a sign that says "I want to pick flowers")

17: 小蜗牛举着牌子，上面写着“我要爬行”(A small snail holding a sign that says "I want to crawl")

18: 小鹿举着牌子，上面显示“我要奔跑”(A little deer holding a sign that says "I want to run")

19: 小猪举着牌子，上面写着“我要睡觉”(A piggy holding a sign that says "I'm going to sleep")

20: 一只机器人在讲台写着“机器学习”(A robot writes "Machine Learning" on a podium)

21: 在游戏大厅中，游戏控制台显示“游戏结束”(In the game lobby, the game console displays "Game Over")

22: 在书店里，一本书被标记为“纽约时报畅销书”(In a bookstore, a book is marked as a "New York Times Bestseller")

23: 小浣熊手里拿着一张写着“我爱阅读”的书签(Little raccoon holding a bookmark that says "I love to read")

24: 课堂上，黑板上写着“知识改变命运”这句话(In class, the sentence "knowledge
changes fate" is written on the blackboard)

25: 在展览馆里，一张写着“请勿触摸”的标牌(In the exhibition hall, a sign says "Do Not Touch")

26: 在医院里，一张写着“请勿打扰”的牌子(In a hospital, a sign that says "Do Not Disturb")

27: 在公共场所，一张写着“禁止吸烟”的标志(A sign that says "No Smoking" in a public place)

28: 在火车站，一张写着“请排队购票”的牌子(At the railway station, there is a sign saying "Please queue up to buy tickets")

29: 在游泳馆，一张写着“请勿奔跑”的标志(At the swimming pool, a sign that says "Do not run")

30: 在工厂里，一张写着“安全第一”的标语(In the factory, a sign that reads "Safety First")

31: 校园里，一块写着“禁止乱扔垃圾”的标语(On campus, a sign reads "No Littering")

32: 在景区里，一张写着“请勿践踏草坪”的标识(In the scenic spot, a sign that reads "Do not trample on the lawn")

33: 在图书馆，一张写着“请勿喧哗”的标牌(In the library, there is a sign
that says "No Noise")

34: 在游泳馆，一张写着“请勿携带食品”的标记(In the swimming pool, a sign that reads "Do not bring food")

35: 在博物馆里，一块写着“禁止用闪光灯”的标牌(In a museum, a sign that reads "No Flash")

36: 在火车站，一张写着“请勿逆行”的牌子(At the train station, a sign that says "Do not go in reverse")

37: 在超市里，一张写着“请勿触摸”的标识(In a supermarket, a sign that says "Do not touch")

38: 在机场，一张写着“禁止携带危险品”的标志(At the airport, a sign that says "Dangerous goods are not allowed")

39: 在停车场里，一张写着“禁止停车”的牌子(In the parking lot, a sign says "No Parking")

40: 一个小姑娘手捧着一本写着“童话”的书(A little girl is holding a book with the words "Fairy Tales" in her hands)

41: 一位摄影师穿着印有“镜头”字样的T恤(A photographer wears a t-shirt with the word "Lens" printed on it)

42: 一名艺术家手握画笔，在画布上勾勒出一幅印有“自由”字样的画作(An artist holds a paintbrush and outlines a painting with the word "Freedom" on the canvas)

43: 印有“梦想”字样的T恤(T-shirt with the word "Dream" printed on it)

44: 印有“科学”字样的书(Books with the word "Science" printed on them)

45: 一名教练拿着印有“努力”字样的牌子(A trainer holds a sign that reads "Effort")

46: 佩戴着印有“自由行”字样的帽子(Wearing a hat with the words "Freedom to Travel" printed on it)

47: 一位志愿者背着印有“爱心”字样的背包(A volunteer is carrying a backpack with the words "Love" printed on it)

48: 一名科学家手持印有“探索”字样的实验器材(A scientist holds experimental equipment with the word "Explore" printed on it)

49: 一辆车上写着“速度”的标语(A car has a slogan with "speed" written on it)

50: 这个盒子上写着“谨慎处理”的警告(The box has a "handle with caution" warning)

51: 这个手机壳上写着“爱护眼睛”的提示(This phone case has reminders to "care for your eyes")

52: 这个包装盒上写着“天然无添加”的字样(It says "Natural No Additives" on the box)

53: 这个旅行箱上写着“轻松出行”的字样(The words "Easy Travel" are written on this suitcase)

54: 这本书上写着“智慧启迪”的标语(The book has a slogan "Wisdom and Enlightenment" written on it)

55: 这个水杯上写着“健康饮水”的标语(This drinking glass has the slogan "Drinking Water Healthy" written on it)

56: 这个化妆品瓶上写着“自然无害”的字样(This cosmetic bottle says 'Natural and Harmless')

57: 这个保温杯上写着“温暖陪伴”的标语(This thermos has the slogan "warm companionship" written on it)

58: 这个垃圾桶上写着“环境保护”的字样(The words "environmental protection" are written on this trash can)

59: 饭盒上写了“健康饮食”的字样(The words "healthy eating" were written on the lunch box)

60: 雨伞上写了“防风防雨”的标语(The slogan "Windproof and rainproof" is written on the umbrella)

61: 水龙头上写了“请勿浪费水资源”的提醒(The reminder of "Don't waste water resources" is written on the faucet)

62: 电池上写了“请勿随意扔弃”的提示("Do not throw away" is written on the battery)

63: 纸巾上写了“方便携带”的字样(The words "easy to carry" are written on the paper towel)

64: 一个红包，写了“新春快乐”的祝福(A red envelope with the blessing of "Happy Chinese New Year")

65: 书包上写了“努力学习”的标语(The slogan "study hard" is written on the schoolbag)

66: 墙壁上写了“严禁乱贴乱画”的规定(On the wall was written the rule of "Strictly No Pasting and Graffiti")

67: 门牌上写了“欢迎光临”的字样("Welcome" written on the door)

68: 这个凳子上写着“不要随地吐痰”的标语(This stool has the sign "Don't Spit" written on it)

69: 婴儿车上贴了“谨防遗失”的提示("Be careful not to lose" reminders posted on baby carriages)

70: 街道上立着“文明交通”的指示牌(There are signs of "civilized traffic" on the street)

71: 电影院门口挂着“禁止使用手机”的告示(A sign saying "No use of mobile phones" is hung at the entrance of the movie theater)

72: 矿泉水瓶上印了“请勿乱扔”的指示("Do not litter" instructions printed on mineral water bottles)

73: 商场里放着“请勿私自拆封”的提示牌(There is a sign saying "Do not unpack without permission" in the mall)

74: 公园里挂着“禁止野餐”的警示牌("No Picnics" signs hang in the park)

75: 餐厅里贴着“请勿随地吐逆”的提醒("Do not spit anywhere" reminder posted in the restaurant)

76: 超市里写着“请勿打开试吃”的提示(A notice saying "Do not open and try" in the supermarket)

77: 书店里贴了“请勿涂鸦”的告示("No graffiti" notice posted in the bookstore)

78: 公共汽车上贴了“让座”的提醒("Give up your seat" reminder posted on the bus)

79: 宾馆里放着“请勿吸烟”的标牌(A "No Smoking" sign is placed in the hotel)

80: 游泳池边挂着“请勿穿鞋入池”的指示(The sign "Do not wear shoes in the pool" hangs beside the swimming pool)

81: 火车站旁架着“乘车注意事项”的告示牌(There is a notice board next to the train station that reads "Travel Precautions")

82: 酒店里贴着“请勿打扰”的门牌(There is a "Do Not Disturb" sign in the hotel)

83: 草地上贴着“禁止踩踏”的标识("No trampling" signs posted on the grass)

84: 电梯里放着“请勿乱按楼层”的告示(There is a sign "Do not press the floor indiscriminately" in the elevator)

85: 家居装饰里挂着“保护环境”的标语("Save the environment" sign with home decor)

86: 超市里放着“八个包装不打折”的告示(There is a notice in the supermarket that says "No discount for eight packages")

87: 椅子上贴了“请勿留座”的提示("Do not reserve a seat" reminder posted on the chair)

88: 医院里挂着“文明就医”的标语(The slogan "Take care in a civilized manner" hangs in the hospital)

89: 饮料瓶上印了“请勿吸管”的提醒("No Straws" warning printed on beverage bottles)

90: 体育馆里放着“请勿在场馆内吸烟”的告示("No Smoking in the Stadium" sign in the gymnasium)

91: 校车上印了“安全第一”的标语("Safety First" slogan printed on school bus)

92: 公园里立着“宜静不宜吵”的标志(In the park, there is a sign "It is better to be quiet than to be noisy")

93: 水族馆里写着“请勿喂食”的提示(A sign saying "Do not feed" in the aquarium)

94: 餐厅里放着“如有不满请及时提出”的告示(There is a notice "If you have any dissatisfaction, please report it in time" in the restaurant)

95: 北极熊手中的牌子写着“请保护我”(A sign in a polar bear's hand reads "Please protect me")

96: 超市里放着“仅限当日购买”的促销海报("Only buy this day" promotional poster in a supermarket)

97: 图书馆里挂着“勿扰”的门牌(There is a "Do Not Disturb" sign in the library)

98: 学校里展示着“廉洁诚信”的口号(The slogan "Integrity and Honesty" is displayed in the school)

99: 公交车站上贴着“乘车文明出行”的标语(The slogan "Travel in a civilized way" is posted on the bus stop)

100: 高速公路上立着“远离疲劳”的提醒牌(Reminder signs of "stay away from fatigue" are erected on the expressway)

101: 电视机上贴了“享受健康生活”的告示("Enjoy a healthy life" notice posted on the TV)

102: 赌场门口挂着“拒绝赌博”的禁止牌(Prohibition sign "No Gambling" hung on the entrance of the casino)

103: 彩票站里写着“理性购彩”的标语(The slogan of "purchasing lottery rationally" is written in the lottery station)

104: 公安局门口印着“严惩不贷”的警示("Severe Punishment" warning printed on the entrance of the Public Security Bureau)

105: 电影院里放着“请勿大声喧哗”的提示("Don't make too much noise" signs in movie theaters)

106: 医院里写着“卫生是每个人的责任”口号(The slogan "Hygiene is Everyone's Responsibility" is written in the hospital)

107: 火车站上贴着“保持环境整洁”的告示牌("Keep your environment tidy" sign posted on the train station)

108: 放映厅里播放着“文明看电影”的宣传片(The promotional video of "Watching Movies in Civilization" is played in the screening hall)

109: 高尔夫球场上挂着“比赛公平竞技”的标语("Play Fair Play" sign on golf course)

110: 一位机器人讲师在黑板上用草书写下“表征学习”字样，并附上数学公式和图表的照片(A robotic lecturer writes the words "representational learning" in cursive on a blackboard, along with a photo of mathematical formulas and diagrams)

111: 一个写着“禁止养狗”但带着狗微笑的标志(a sign that reads “no dogs” but with a dog smiling)

112: 一个用粗体字母写着“行星地球”的地球仪，各大洲用鲜艳的颜色表示(A globe with the words "Planet Earth" written in bold letters, with continents in bright colors)

113: 一张四周都是玫瑰花的照片，远处有一个牌子，上面写着“危险雷区”(A photo of roses surrounded by a sign in the distance that reads "Danger Minefield")

114: 木制长颈鹿牙刷，彩虹色的“长颈鹿牙刷”字样(Wooden Giraffe Toothbrush with "Giraffe Toothbrush" lettering in rainbow colors)

115: 一架飞机在城市上空飞行，烟雾轨迹上写着“支持”(A plane flies over the city with the words "support" written in smoke trails)

116: 一张熊猫在一个大会议室里做演讲的照片，上面写着“扩散模型”(A photo of a panda giving a presentation in a large conference room, with text "Diffusion Models")

117: 两只美洲驼在曼波舞，指着一块写着“曼波”的牌子(Two llamas doing a mambo dance and pointing at a sign that says "Mambo")

118: 一个手绘的菠萝形状的木制“菠萝俱乐部”标志，悬挂在酒吧外(A hand painted wooden "Pineapple Club" sign in the shape of a pineapple, hanging outside a bar)

119: “空灵媒体”公司的标志，其中字母看起来像一幅画("Ethereal Media"'s logo, where the letters look like a painting)

120: 《融化的雪人》乐队专辑“神秘的插曲”的封面(Cover of Melting Snowman's album "Mysterious Interlude")

121: 显示“优化模式”字样的拼字板(Scrabble board showing the words "optimized mode")

122: 美丽花园里的花朵，花朵写着“和平”(Flowers in a beautiful garden with the word "Peace" written)

123: 一张详细的图纸，文字为“复古刻字”，字母主义，粗规格花丝(A detailed drawing with the text "vintage lettering", alphabetism, thick gauge filigree)

124: 香蕉放在野餐桌上，形成“那是香蕉”(Bananas placed on a picnic table forming "That's a Banana")

125: 一个标有“能量补品”的古董瓶(An antique bottle labeled "Energy Tonic")

126: 一架直升机侧面写着“直升机之旅”的照片，降落在山谷中的直升机停机坪上，背景是河流、树木和山脉(Photo of a helicopter with "helicopter tour" written on the side, landing on a helipad in a valley with river, trees and mountains in the background)

127: 带有“单向”标志的照片(photo of a sign with “one way”)

128: 一个由金属丝和纸制成的大脑雕塑，在大脑材料中写着“深刻的思想”(A brain sculpture made of wire and paper with "deep thoughts" written in brain material)

129: 一家连锁杂货店的标志，名称为“杂货店”(Logo for a chain of grocery stores with the name "Grocery")

130: 由彩色细线制成的文字“解锁创造力”雕塑摄影棚(Text "Unlock Creativity" sculpture photo booth made of thin colored lines)

131: 工作室拍摄的一双由彩色电线制成的鞋子雕塑和文字“解锁创造力”(Studio shot of a pair of shoe sculptures made from colored wires and the text "Unlock Creativity")

132: 拉斯维加斯大道的一幅复古图片，用粗体印刷字体写着“拉斯维加斯”(A vintage image of the Las Vegas Strip with the words "Las Vegas" in bold print)

133: 一个机器人在黑板上用粉笔写下“道德”(A robot writes "morality" in chalk on a blackboard)

134: 彩虹色烟雾中的黄色萨克斯风，上面写着“时髦的烟雾”，看起来像音乐般的烟雾(A yellow saxophone in rainbow colored smoke with the words "Funky Smoke" looking like musical smoke)

135:一本题为“知识就是力量”的古书的摄影棚特写(studio close-up shot of an antique book with "knowledge is power")

136: 一只鹦鹉的肖像正举着一个牌子，上面写着“演示文稿时”(A portrait of a parrot is holding a sign that says "During presentations")

137: 一张蒲公英田的照片，上面写着“草坪修剪时，蒲公英是第一个离开的”(A photo of a dandelion field with the caption "When the lawn is mowed, the dandelions are the first to leave")

138: 金叶曼荼罗中心的泰姬陵，底部有“荣誉之地”字样(The Taj Mahal in the center of a gold leaf mandala with the words "Place of Honor" at the bottom)

139: 一张名为“北美鹌鹑”的海报，展示了不同种类的鹌鹑(A poster titled "North American Quail" showing different species of quail)

140: 一只猫的卡通画，上面有一个思想泡泡，说“这太奇怪了”(A cartoon of a cat with a thought bubble that says "this is weird")

141: 海盗船上的一只鹦鹉，一只鹦鹉戴着海盗帽，并配文“我现在是船长了”(A parrot on a pirate ship, a parrot wearing a pirate hat with the text "I'm the captain now")

142: “时间是暂时的一切都是暂时的”的生成艺术，由点、河流、图形设计和白色背景构成的粘稠烟雾("Time is temporary and everything is temporary" generative art, sticky smoke made of dots, rivers, graphic design and white background)

143: 用热狗、博物馆品质、相框照片和白色背景制作的“食物太糟糕了分量太小”字样的摄影棚(Studio shot of "Food Too bad the portions are too small" made of hot dogs, museum quality, framed photo and white background)

144: 一张看起来很强大的汽车图片，看起来像是为越野而设计的，文字上写着“我是卡车不是汽车”(An image of a powerful looking car that looks like it was built for off-roading with the text "I'm a truck not a car")

145: 一个森林的极简主义版本，前面有一个“帮助森林”的标志(A minimalist version of a forest with a "Help the Forest" sign on the front)

146: 一只戴着厨师帽的狗的漫画，带有一个思想泡泡，上面写着“我什么都记不起来了”(A cartoon of a dog wearing a chef's hat with a thought bubble that says "I can't remember anything")

147: 一个复古咖啡广告，上面写着“咖啡是我喜欢的”(A retro coffee ad that says "coffee is my thing")

148: 公园长椅一端的风景，望着天空，天空中写着“想象结果”(View from one end of a park bench, looking at the sky where "Imagine Result" is written)

149: 一只巨大的鞋子，上面写着“做作的鞋子”(a giant shoe, with the caption "shoe for hokey pokey")

150: 一份报纸的标题是“本地猪吃奖品南瓜”，还有一张照片显示了吃了一半的南瓜(A newspaper headline read "Local pig eats prize pumpkin" and a photo showed a half-eaten pumpkin)

151: 一个写着“世界上最好的熟食店”的店面，居中(A storefront that says "World's Best Deli", centered)

152: 一个盘子，里面有一个牡蛎，牡蛎上有一把叉子和一把刀，上面写着“午餐牡蛎”(A plate with an oyster in it, a fork and a knife on the oyster and the words "LUNCH OYSTER")

153: 机器人正举着一个写着“我不是机器人”的牌子(A robot is holding a sign that says "I'm not a robot")

154: 摄影棚拍摄的文字“我喜欢咖啡因为它给我一种可能醒着的错觉”由咖啡液制成，博物馆品质，白色背景(Studio shot of text "I like coffee because it gives me the illusion that I might be awake" made of liquid coffee, museum quality, white background)

155: 一张匆忙手写的纸条，上面写着“我四点回来”，贴在冰箱上(A hastily handwritten note saying "I'll be back at four" posted on the fridge)

156: 一本名为“秘鲁食谱”的大型食谱书(A large recipe book called "The Peruvian Cookbook")

157: 带有“我对移动楼梯的恐惧正在升级”的广告牌(Billboard with "My fear of moving stairs is escalating")

158: 一块石头的影子，从一只蚂蚁的角度拍摄，标题是“看看那个影子”(The shadow of a rock, photographed from the perspective of an ant, captioned "Look at that shadow")

159: 一个南瓜，留着胡子，戴着单眼眼镜，头戴礼帽，在一个演讲泡泡里写着“你也能致富”(A pumpkin with a beard, a monocle and a top hat with the text "You Can Get Rich too" in a speech bubble)

160: 一幅漫画，描绘了一只狗拿着望远镜看着一颗带有讲话气泡的恒星，上面写着“我想知道那个星球上有没有狗”(A cartoon of a dog holding a telescope looking at a star with a speech bubble that says "I wonder if there are dogs on that planet")

161: 房子的蓝图，屋顶为三角形，墙壁为正方形，地板为矩形，并带有“这座房子是基于抽象原则建造的”的信息(Blueprint of a house with a triangular roof, square walls and rectangular floor with the message "This house is built on abstract principles")

162: 一块向日葵地，拖拉机即将碾过向日葵，标题是“向日葵过后，他们会来找你”(A field of sunflowers, a tractor is about to run over the sunflowers, with the caption "After the sunflowers pass, they will come to you")

163: 文字“气球正在飞行”，由彩虹气球制成，彩色背景(Text "Balloons are flying", made of rainbow balloons, colorful background)

164: 哈勃望远镜和银河系，文字为“宇宙是一个谜但我们在这里解决它”(Hubble and the Milky Way with the text "The universe is a mystery but we are here to solve it")

165: 一颗写着“我爱你”的心，用彩虹色写着(A heart with the words "I love you" written in rainbow colors)

166: 白色背景、标题为“如何成为管理者中的管理者”的美丽教科书的摄影棚照片(Studio shot of a beautiful textbook with the title "How to be a Manager of Managers" on a white background)

167: 一张装饰性贺卡，上面写着“恭喜你达到了艺术水平”(A decorative greeting card that reads "Congratulations on reaching the level of art")

168: 一个牌子上写着“请不要和黑猩猩争论”(A sign reads "Please don't argue with chimpanzees")

169: 一幅乌龟头上有一个思想泡泡的漫画，上面写着“如果没有思想泡泡怎么办”(A cartoon of a turtle with a thought bubble on its head and the words "what if there is no thought bubble")

170: “秋天来了”写在漂浮在湖面上的秋叶上("Fall is here" written in autumn leaves floating on a lake)

171: 一只螃蟹坐在沙滩上，拿着冲浪板，太阳是一个巨大的橙色，天空是一道彩虹，螃蟹在想“你才是最重要的”(A crab is sitting on the beach with a surfboard, the sun is a huge orange and the sky is a rainbow, and the crab is thinking "you are the most important thing")

172: 从飞机上看到的多伦多市，画面中央有一座巨大的塔，漫画中有文字“巨塔”(The city of Toronto seen from the plane, with a huge tower in the center of the frame, with the text "Giant Tower" in the cartoon)

173: 一幅河马的漫画，上面有一个演讲泡泡，说“我是河马你想要什么”(A cartoon of a hippopotamus with a speech bubble saying "I am a hippopotamus what do you want")

174: 一只穿着西装打着领带的龙虾，拿着麦克风，上面写着“龙虾说什么”(A lobster in a suit and tie holding a microphone with the words "What does the lobster say")

175: 一本以“手术变得容易”为标题的书(A book titled "Surgery Made Easy")

176: 椅子的艺术装置，靠背上刻着“我一无所获”(An art installation of a chair with "I got nothing" engraved on the back)

177: 一幅风景画，上面写着“这幅画不是我画的”(A landscape painting with the words "I didn't paint this picture")

178: 一张苹果擦伤的照片，用花哨的字体写着“苹果对你有益”(A photo of a bruised apple with the words "apples are good for you" written in fancy lettering)

179: 一张柯基犬的照片，上面写着“我不是真正的柯基犬”(A picture of a corgi that says "I'm not a real corgi")

180: 单词“需要人工智能和雨水才能形成彩虹”黑色背景，全息照相，霓虹色(Words "It takes artificial intelligence and rain to make a rainbow" on black background, holographic, neon colors)

181: “每个艺术家最初都是业余爱好者”字样上的黑白标志白色背景、线框图、生成艺术(Black and white sign with the words "Every artist is first an amateur" on a white background, wireframe, generative art)

182: 两只手的照片，一只手抱着一颗心，另一只手拿着一个闪电，上面写着“爱就是力量”(Photo of two hands holding a heart in one and a lightning bolt with the words "Love is power")

183: 阿尔卑斯山的美丽照片，标题为“最美山脉”(Beautiful photo of the Alps with the caption "Most beautiful mountains")

184: 一棵树的铅笔素描，标题是“这里没有树”(A pencil drawing of a tree with the caption "There are no trees here")

185: 一片黑暗的森林，远处只有一盏灯，文字“我又来和你说话了”(A dark forest with only one light in the distance and the text "I've come to talk to you again")

186: 一个圆圈，上面写着“无限让我快乐”，字体看起来像是手写的(A circle with the words "Infinity makes me happy" written in lettering that looks like it was handwritten)

187: 以文本“知识就是力量”萌芽、居中的形式拍摄藤蔓(A shot of a vine with the text "Knowledge is power" sprouting, centered)

188: 一张美丽罂粟田的照片，上面写着“请勿拍照”(A photo of a beautiful poppy field with the words "Do not take pictures")

189: 一株带有“无太阳能电池板”标志的暴躁向日葵(A cranky sunflower with a
“No Solar Panels” sign)

190: 一个试管，里面有一滴液体，上面写着“我们在火星上发现了水”(A test tube with a drop of liquid inside and the words "We found water on Mars")

191: 背景是一座城市，前景是一片云彩，用圆形草书写着“沉思云彩”(With a city in the background and a cloud in the foreground, the word "Contemplative Cloud" written in round cursive)

192: 一只狗和一只猫的照片，它们的头从笼子里伸出来，上面写着“禁止养宠物”(Photo of a dog and a cat with their heads sticking out of a cage with the words "No pets allowed")

193: 一台20世纪80年代风格的电脑的三维模型，屏幕上写着“我的旧习惯”(a 3d model of a 1980s-style computer with the text “my old habit” on the screen)

194: 一只拿着手电筒的老鼠说“我害怕黑暗”(A mouse with a flashlight says "I'm afraid of the dark")

195: 一张兔子喝咖啡看书的照片书名“彼得兔历险记”可见(A photo of a rabbit drinking coffee and reading a book titled "The Adventures of Peter Rabbit" is visible)

196: 小丑拿着一张纸牌，上面写着“即使在艰难时期也有可能玩得开心”(The clown holds a card that says "It's possible to have fun even when times are tough")

197: 报纸，标题为“在太空中发现的外星人”(Newspaper with the headline "Aliens Found in Space")

198: 一只狗有一个带着“汪汪其他狗会烦我们”字样的翻译演讲气泡(A dog has a translated speech bubble with the words "woof woof other dogs annoy us")

199: 黄油食品加工线上的机器人，机器人神情沮丧，头顶上的红灯指示错误，机器人说“我不敢相信这不是黄油”(Robot on butter food processing line, robot looking frustrated, red light above head indicates wrong, robot says "I can't believe it's not butter")

200: 墙上“释放粉色”的涂鸦艺术("Release the Pink" graffiti art on the wall)

201: 一只蜥蜴坐在棒球场的本垒板上，在一个演讲气泡中写着“使其安全”(A lizard sitting on home plate at a baseball field with the words "Make it safe" in a speech bubble)

202: 一张处于不同发展阶段的多棵树的照片，并配文“生长是一个持续的过程”(A photograph of multiple trees in various stages of development with the caption "Growing is an ongoing process")

203: 一朵紫色的花，头上戴着皇冠，还有一个演讲泡泡，上面写着“我是紫色的花”(A purple flower with a crown on her head and a speech bubble that says "I am a purple flower")

204: 一个50年代风格的机器人，头部巨大，身体形状像火箭，并配文“一个真正的太空人”(A 50s-style robot with a huge head and a body shaped like a rocket, with the caption "A real spaceman")

205: 一家名为“为你所需”的面包店的专业设计标志(Professionally designed logo for a bakery called "All You Need")

206: “这是未来”一词的最小雕塑，由浅金属彩虹色铬细线制成，3D渲染，等距透视，超详细，深色背景(Minimal sculpture of the word "This is the future", made of light metallic iridescent chrome thin lines, 3D rendering, isometric perspective, super detailed, dark background)

207: 枕头形状为“周末准备好了”，字母主义，有趣的杂乱字母，[特写]面包，作者不详，平面艺术(Pillow in the shape of "Weekend Ready", Alphabetism, fun jumbled letters, [close-up] bread, author unknown, graphic art)

208: 在一个有“请勿触摸”标志的漂亮花盆中的植物(plant in a fancy pot with a
"do not touch" sign on it)

209: 一张画着“拯救地球”的地球图片(A picture of the Earth with the words "Save the Planet")

210: 学者大象阅读一份标题为“大象统治世界”的报纸(Scholar Elephant reads a newspaper headlined "Elephants Rule the World")

211: 一张写着“在海滩上养一只名叫鲨鱼的狗是个错误”的标牌照片(A photo of a sign that reads "Having a dog named Shark on the beach was a mistake")

212: 地球被合并的多个闪电击中的照片插图，标题为“对光速的惊讶”(Photo illustration of Earth being struck by multiple lightning bolts merging, titled "Amazing at the Speed of Light")

213: 一张里面有鱼的鱼缸照片，上面写着“鱼缸你来参观”(A photo of an aquarium with fish in it, with the words "Fishbowl you visit")

214: 白色背景上的颜料飞溅中的“艺术永远不会结束只有继续”，涂鸦艺术，虚无的边缘，爱，泥泞的颜色，彩色木刻，美丽，光谱色("Art never ends only goes on" in paint splatter on white background, graffiti art, edge of nothingness, love, muddy colors, colorful woodcut, beautiful, spectral colors)

215: 餐馆“加油站”的照片(Photo of the restaurant "Gas Station")

216: 一件t恤，上面写着“没有行星”(A t-shirt with the message
"There is no planet B" written on it)

217: 一个牙膏管小雕像的特写，3D渲染，糖果粉彩，管上有“刷牙”的文字(Close up of a toothpaste tube figurine, 3D rendering, candy pastels, with the text "Brush your teeth" on the tube)

218: 一张手工绘制的时光机蓝图，标题为“时光旅行设备”(A hand-drawn blueprint of a time machine titled "A Time Travel Device")

219: 在音乐会的现场，“摇滚”字样(At the concert site, the words "rock")

220: 图书馆墙壁上贴着“请勿涂鸦”的标语(There are slogans on the walls of the library that say "No Graffiti")

\end{CJK}


\end{document}